\documentclass[]{siamltex}
\usepackage[utf8]{inputenc}
\usepackage[british]{babel}
\usepackage[mono=false]{libertine}
\usepackage[T1]{fontenc}
\usepackage[top=2.5cm, bottom=2.5cm, left=2cm, right=2cm]{geometry}
\usepackage[font=small]{caption}
\usepackage{subcaption}
\usepackage{cite}
\usepackage{verbatim}
\usepackage{enumitem}
\usepackage{titlesec}
\usepackage[usenames, dvipsnames]{color}
\usepackage[titletoc]{appendix}
\usepackage[pdftex]{graphicx}
\usepackage{array}
\usepackage{url}
\usepackage{mathtools}
\usepackage{amssymb,amsfonts,amsmath}
\usepackage{dsfont}
\usepackage{bm}

\usepackage{amssymb}
\usepackage{amsfonts}
\usepackage{amssymb}
\usepackage{graphicx}
\usepackage{comment}

\newcommand{\rhD}{\hat\rho_h^{\mathcal{D}}}

\newcommand{\ve}{{\varepsilon}}
\newcommand{\ove}{\overline{\varepsilon}}
\newcommand{\og}{\overline{g}}
\newcommand{\cS}{{\mathcal{S}}}
\newcommand{\cD}{{\mathcal{D}}}
\begin{document}
\setcounter{tocdepth}{2}
\setcounter{secnumdepth}{3}
\title
{Kernel Density Estimators in Large Dimensions}
\author{Giulio Biroli\textsuperscript{1} and Marc M\'ezard\textsuperscript{2}}
\date{}
\maketitle

\begin{center}
$^1$ Laboratoire de Physique de l'Ecole Normale Sup\'erieure, ENS, Universit\'e PSL, CNRS, Sorbonne Universit\'e, Universit\'e Paris-Diderot, Sorbonne Paris Cit\'e, Paris, France\\ 
$^2$ Department of Computing Sciences, Bocconi University
\end{center}

\begin{abstract}
    This paper studies Kernel Density Estimation for a high-dimensional distribution $\rho(x)$.  Traditional approaches have focused on the limit of large number of data points $n$ and fixed dimension $d$. We analyze instead the regime where both the number $n$ of data points $y_i$
and their dimensionality $d$ grow with a fixed ratio $\alpha=(\log n)/d$. 
 Our study reveals three distinct statistical regimes for the kernel-based estimate of the density $\hat \rho_h^{\mathcal {D}}(x)=\frac{1}{n h^d}\sum_{i=1}^n K\left(\frac{x-y_i}{h}\right)$, depending on the bandwidth $h$: a classical regime for large bandwidth where the Central Limit Theorem (CLT) holds, which is akin to the one found in traditional approaches. 
 Below a certain value of the bandwidth, $h_{CLT}(\alpha)$, we find that the CLT breaks down. The statistics of $\rhD(x)$ for a fixed $x$ drawn from $\rho(x)$ is given by a heavy-tailed distribution (an alpha-stable distribution). In particular below a value $h_G(\alpha)$, we find that $\rhD(x)$ is governed by extreme value statistics: only a few points in the database matter and give the dominant contribution to the density estimator. We provide a detailed analysis for high-dimensional multivariate Gaussian data. We show that the optimal bandwidth threshold based on Kullback-Leibler divergence lies in the new statistical regime identified in this paper. As known by practitioners, when decreasing the bandwidth a Kernel-estimated estimated changes from a smooth curve to a collections of peaks centred on the data points. Our findings reveal that this general phenomenon is related to sharp transitions between phases characterized by different statistical properties, and offer new insights for Kernel density estimation in high-dimensional settings.
\end{abstract}

\section{Introduction}
Given a data set, a standard problem in statistics is to estimate the density of probability from which it has been generated. This problem has been widely discussed in the case where the data points belong to a space with few dimensions. In this paper we shall discuss the case where the data points belong to a large-dimensional space. This is particularly relevant for modern developments of artificial intelligence. For instance, generative modelling with diffusion or flows \cite{Sohl_Dickstein2015,song2020denoising} consists in generating new points from an unknown underlying probability, given a database of examples. It thus amounts to estimating the probability density, with enough precision so that one can generate new examples from this unknown probability. Moreover, the density generated by  diffusion models has indeed the form of a Kernel-estimated density when the exact empirical score is used. Many examples of such generation have been proposed in recent years, ranging from images, videos, or scientific data such as turbulent flows\cite{Sohl_Dickstein2015,Song2019,Song_Sohl-Dickstein2021,guth2022wavelet,yang2022diffusion,saharia2022photorealistic,bar2024lumiere,poole2022dreamfusion}. In all these cases, data generally live in a large dimensional space, and we shall argue below that standard methods for density estimation do not apply in this limit. 

Let us be more specific. Given $n$ data points in $\mathbb{R}^d$, drawn independently from an
unknown distribution with density $\rho$, a standard
method to estimate this distribution uses a positive density kernel $K(x)$ to construct the estimator of the density at point $x$:
\begin{align}
\hat \rho_h^{\mathcal {D}}(x)=\frac{1}{n h^d}\sum_{i=1}^n K\left(\frac{x-y_i}{h}\right)
\label{rhD_def}
\end{align}
where ${\mathcal {D}}= \{y_i\}$, $i\in\{1,...,n\}$ are the data, and $h$ is a bandwidth parameter which must be 
optimized.

It is useful to first summarize the usual way to find the optimal kernel bandwidth $h$ in finite dimension $d$ \cite{wasserman2006all} for large $n$.  This is obtained
by minimizing the average mean square error
\begin{align}
{\mathcal {L}}_2= {\mathbb {E}}_\mathcal {D}\; \int d^dx\; \left[\rho(x)-\hat \rho_h^{\mathcal {D}}(x)\right]^2
\end{align}
where  ${\mathbb {E}}_\mathcal {D}$ is the empirical average with
respect to the database, and $n$ is assumed large enough to apply the central limit theorem. This minimization involves a balance between
bias and variance. When the density probability law $\rho(x)$ is
regular enough on the scale $h$, one can expand the bias at small $h$, and the result is the Scott and Wand formula \cite{scott1991feasibility}:
\begin{align}\label{eq:scottwand}
  \rhD(x)=\rho(x)+\kappa\frac{h^2}{2}\Delta \rho(x)+
  \frac{1}{\sqrt{n h^d}}\; \sqrt{c_2 \rho(x)}\; z(x)
\end{align}
where $z(x) $ is a gaussian random variable with zero mean and
variance unity, and $c_2=\int dz K(z)^2$.
Substituting this into the mean square error, and assuming a rotation invariant kernel such that $\int d^dx \; K(x)x_i x_j= \kappa \delta_{ij}$, the optimal value of $h$ is found equal to
\begin{align}\label{eq:hstar}
    h^*=n^{-1/(d+4)}\left[\frac{ c_2\; d}{\kappa^2 \int dx\; [\Delta \rho(x)]^2}\right]^{1/(d+4)}
\end{align}

This formula shows a well-known effect named the curse of dimensionality: when $d$ is large, one cannot get  a good approximation of $\rho$ with the kernel, unless the number of data points $n$ scales exponentially with $d$. 
On the other hand this analysis, as most of the statistics literature, focuses on the limit of large $n$ at fixed $d$.


In this work we are interested in density estimation "in large dimensions", which is the regime where both $n$ and $d$ go to infinity, with  $\alpha=\log n /d$  fixed. 
It is well known \cite{wainwright2019} that smooth enough densities can be studied in this limit thanks to some concentration properties.
In fact this will enable us to show that the classic analysis of kernel density estimation then enters into a new statistical regime.  In the next section we illustrate this new regime through a simple example, from which we give a high level overview of our main results. The following sections develop the formal setup, list the results, and give the proofs.

Before moving to the description of this new regime, we would like to underline the fact that the exponential regime $n=e^{\alpha d}$ is not only of theoretical interest, it is also relevant in practice for the scale of database used in machine learning.  For instance, studying images in dimension $d=500$ and a data base of $10,000$ points results in a value of $\alpha=.018$ for which we will see that our approach gives interesting new predictions.

\section{A high-level overview: The three statistical regimes }

In Kernel Density Estimation it is often assumed that $n$ is large enough to be able to apply the central limit theorem (CLT). In this case $\hat \rho_h$ can be written as an average term (the bias) and a Gaussian random part (the variance), see eq. (\ref{eq:scottwand}). In the limit of large dimensions $n=e^{\alpha d}$, even for very large values of $n$, this decomposition in bias and variance does not hold generically: it is correct only when $h>h_{CLT}(\alpha)$, where the critical value of the bandwidth $h_{CLT}(\alpha)$ is an increasing function of $\alpha$. In fact, there exists a new regime at $h<h_{CLT}(\alpha)$, in which the statistics of $\hat \rho_h$ is not governed by the CLT. This new regime  can itself be divided into two phases separated by a critical value $h_G(\alpha)$: for $h
<h_{G}(\alpha)$, a condensation effect typical of glassy phases in physics takes place: the sum defining $\hat \rho_h^{\mathcal{D}}$, although it contains an exponential number of terms, is actually dominated by a finite number of them. In order to find the optimal bandwidth, a suitable criterion  in the large-$d$ limit is to minimize 
the Kullback-Leibler distance. In the cases studied here, the optimal bandwidth is found in the glassy phase 
$h
<h_{G}(\alpha)$.

\subsection{A simple example: the Isotropic Normal case}
To illustrate the main point of our work, let us first focus on a very simple case in which the data is isotropic and Gaussian with mean zero and unit variance,  $\rho(x) \sim {\mathcal N}(0,{\mathcal{I}})$ and the kernel used for density estimation is also Gaussian: 

\begin{align}
\rhD(x)=\frac{1}{n }\sum_{i=1}^n \frac{1}{\sqrt{2\pi  h^2}^d}\exp\left(-\frac{(x-y_i)^2}{2 h^2}\right)
\end{align}
This is a case in which the optimal bandwidth associated to the mean square error can be obtained exactly, see \cite{epanechnikov1969non}. 
For a given $x$, sampled from the distribution $\rho$, the estimator $\rhD$ is a sum of an exponential number of terms, but each term itself scales exponentially with $d$. We shall show that in this regime, the CLT does not always hold. When it holds, $\rhD(x)$ concentrates around its mean $\mathbb{E}  \rhD(x)$. 
\begin{figure}
\centering
    \includegraphics[width=0.6\textwidth]{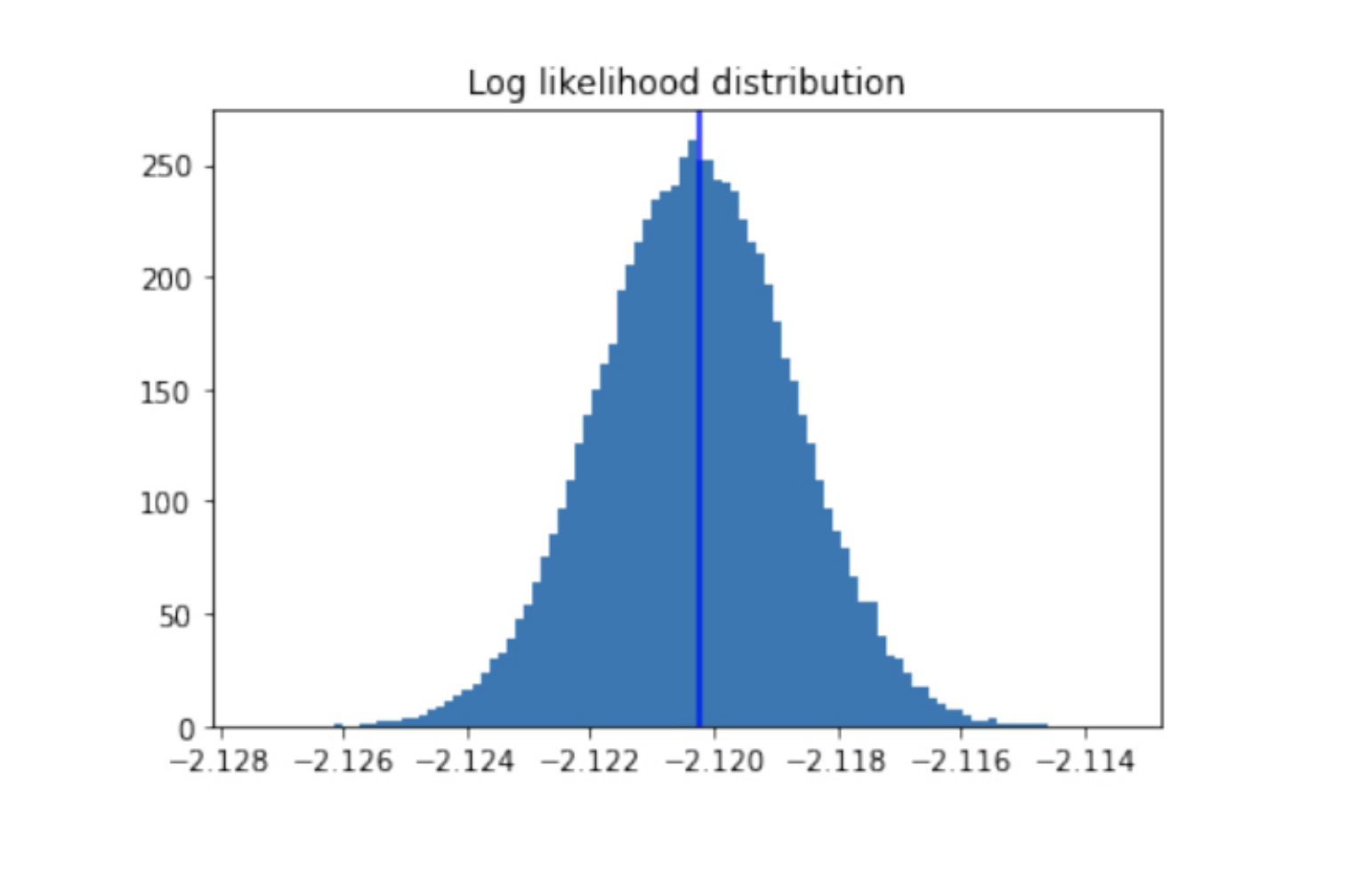}
    \caption{
    Numerical distribution of $\frac{1}{d}\log \rhD(x)$ for for $n=164$, $d=51$ and $h=3$. The distribution is obtained generating $10^5$ samples of the datapoints $\{y_i\}$. The vertical line corresponds to the value of $\frac{1}{d}\log {\mathbb {E}}_{\mathcal{D}}[\rhD(x)]$. 
    }
    \label{fig:clt}
\end{figure}
We study study numerically a case with  $n=164$, $d=51$ (therefore $\alpha=.1$), by drawing one $x$ randomly sampled from $\rho$, and computing numerically the distribution over the database $\mathcal{D}=\{y_i\}$. 

\begin{figure}
    \includegraphics[width=0.5\textwidth]{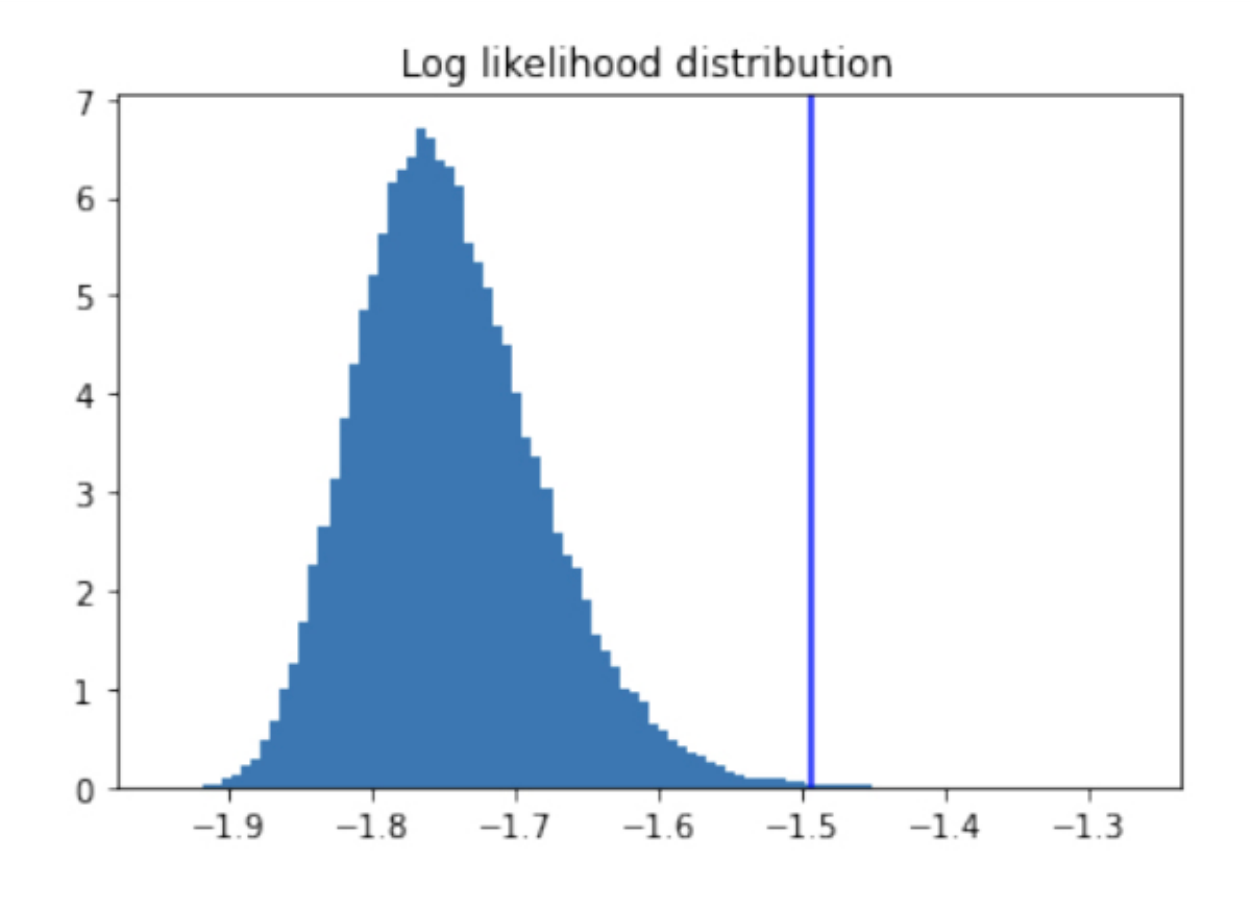}
     \includegraphics[width=0.5\textwidth]{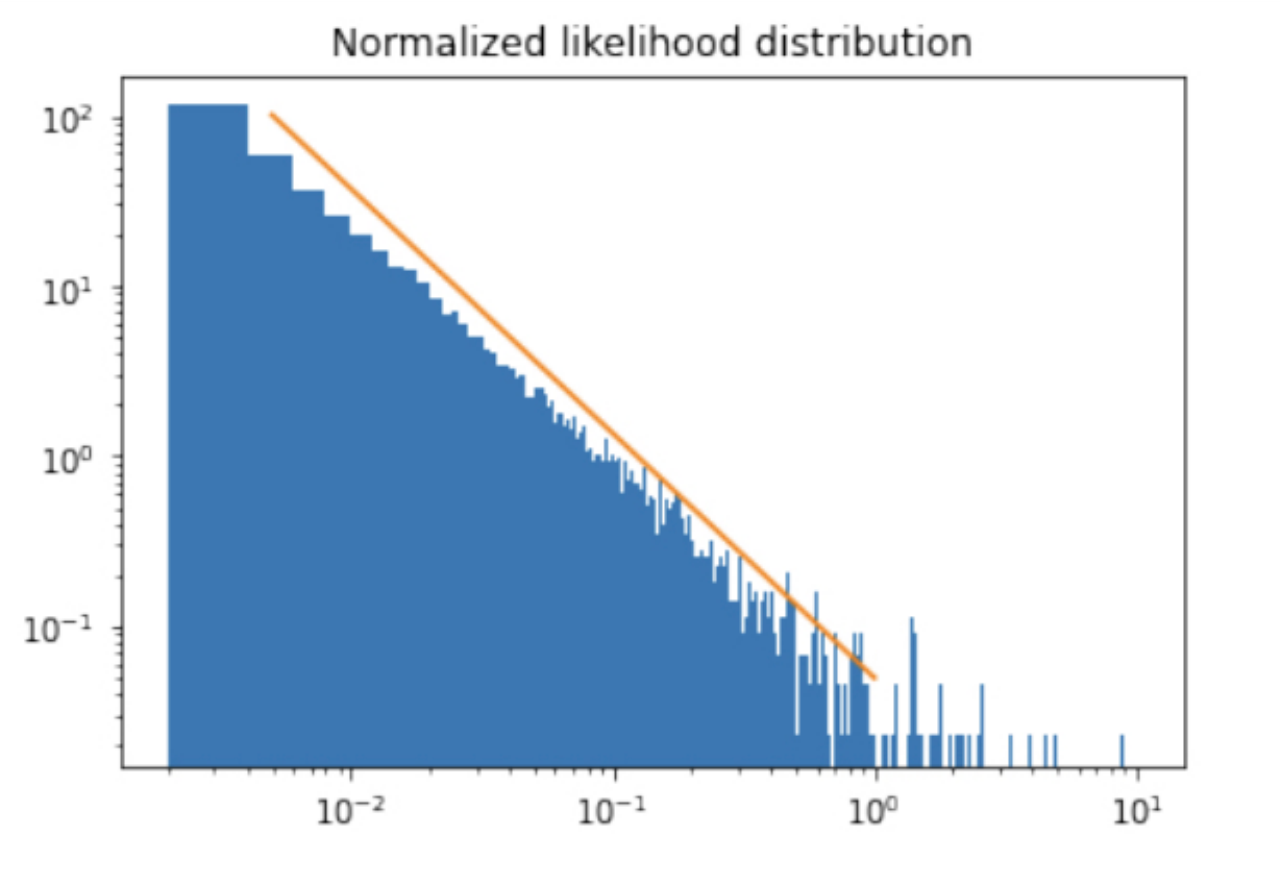}
    \caption{
    Left: Numerical distribution of $\frac{1}{d}\log \rhD(x)$ for for $n=164$, $d=51$ and $h=0.9$. The distribution is obtained generating $10^5$ samples of the datapoints $\{y_i\}$. The vertical line corresponds to the value of $\frac{1}{d}\log {\mathbb {E}}_{x_i}[\rhD(x)]$. Right: Numerical distribution of $z=\rhD(x)/\hat\rho_h^{typ}(x)$ in a log-log scale. The line corresponds to a power law with $z^{-1-m}$ with exponent $m=0.435715$ (obtained from our theory, see Sec. \ref{sec:proba_distrib_Z}). 
    }
    \label{fig:rare}
\end{figure}

In Fig. \ref{fig:clt} we first study the case when $h=3$. As $\rhD$ scales exponentially with $d$ we plot the distribution of
 of $\frac{1}{d}\log \rhD(x)$. We observe that it is peaked around $\frac{1}{d}\log {\mathbb {E}}_{\mathcal{D}}[\rhD(x)]$.  This figure
 shows an example for $n,d\rightarrow \infty$ in which CLT holds. Moreover, one does not need to reach huge values of $n$ to be in the CLT regime, even if $d$ is not small.  
For smaller values of $h$ the situation changes drastically. In Fig \ref{fig:rare} we show the same data  for $h=0.9$ which is a value close to the exact optimal bandwitdth for the mean square error and $n=164$, $d=51$. 
The first striking result of Fig. \ref{fig:rare} (left) is that the distribution of $\frac{1}{d}\log 
\rhD(x) $ is peaked, but it is not centered at $\frac{1}{d}\log {\mathbb {E}}_{\mathcal{D}}[\rhD(x)]$.
The average value of the estimator is not representative of the typical behaviour: it is dominated by rare events of $\rhD(x)$. The distribution of $\rhD(x)$ unveils the reason for this result. We first normalize $\rhD(x)$ with respect to its typical value
defined as $\hat\rho_h^{typ}(x)=\exp \left({\mathbb {E}}_{\mathcal{D}}[\log \rhD (x)]\right)$, and then show its numerical distribution in Fig \ref{fig:rare} (right). The law of the variable $z=\hat\rho_h(x)/\hat\rho_h^{typ}(x)$ is heavy-tailed, behaving at large $z$ proportionally to $ z^{-(1+m)}$ with a value $m<1$.  This is a case in which the CLT does not hold, and hence the usual reasoning based on moments of the kernel, bias (first moment) and variance (second moment), is no longer valid. As we shall show, this regime is generically present for large $n$ and $d$. Our framework will allow to characterize this regime in  detail. 

\subsection{Short overview of our main results.}
  We focus on specific classes of kernels and data which are defined in Secs \ref{def_kernel} and \ref{def_dens}. Our results could be derived in a more general setting, like mixtures of high-dimensional Gaussian distributions and probability distributions associated to statistical physics models (Ising ferromagnets, Hopfield models,...).  
  \subsubsection{Three statistical regimes.}
  In the limit $n,d \rightarrow \infty$ with fixed $\alpha=(\log n)/d$, we show that there exist three statistical regimes for 
$\rhD(x)$ (henceforth we shall always consider that $x$ is fixed and drawn from $\rho(x)$):
\begin{enumerate}
\item For $h>h_{CLT}(\alpha)$,  CLT holds: 
\begin{align}
\frac{\rhD(x)-{\mathbb {E}}_{\mathcal{D}}[\rhD(x)]}{\sqrt{\mathrm{Var}_{\mathcal{D}}[\rhD(x) ]}}\rightarrow g
\end{align}
where $g$ is a Gaussian variable with distribution $\mathcal{N}(0,1)$.  
\item for $h_G(\alpha)<h<h_{CLT}(\alpha)$, the CLT does no longer hold, but the law of large numbers is still valid:
\begin{align}
   z= \frac{\rhD(x)}{{\mathbb {E}}_{\mathcal{D}}[\rhD(x)]}\rightarrow 1 \,.
\end{align}
In this case, the fluctuations of $\rhD(x)$ when one changes the database are not of the order of the square root of the variance,  and they are not Gaussian. The variable $z$  is instead distributed around one following an $\alpha$-stable law which behaves at large $z$ as $c z^{-(1+m)}$ with  $1<m<2$.
\item For $0<h<h_G(\alpha)$,  also the law of large numbers breaks down. While
\begin{align}
   \frac{ \log \rhD(x)}{{\mathbb {E}}_{\mathcal{D}}\log \rhD(x)}\to 1\ ,
\end{align}
one finds that 
\begin{align}
    \frac{\rhD(x)}{\exp({\mathbb {E}}_{\mathcal{D}}[\log \rhD(x)])}\rightarrow z 
\end{align}
where $z$ is  a L\'evy $\alpha$-stable random variable which behaves at large $z$ as $c z^{-(1+m)}$ with  $0<m<1$.  In this regime the average value of $\rhD(x)$ becomes much larger than the typical value. Its large value is due to exponentially rare samples of $\mathcal{D}$ which bias the average. 
\end{enumerate}
The values of $h_G(\alpha)$ and $h_{CLT}(\alpha)$ depend on $\alpha$, and they are monotonously decreasing with $\alpha$.  
These three regimes are, mutatis mutandis, the ones found for the Random Energy Model, which was studied in physics as a simple disordered system\cite{derrida1981random}. 
They have been discussed in some generality for sums of exponentials of random variables  \cite{bovier2002fluctuations,ben2005limit}. The setup which we study here is a generalization of this work to the case where the distribution of the variables have a large deviation description, with a parameter related to the number of variables in the sum. In fact, the Kernel Density Estimation (\ref{rhD_def}) can be rewritten as a partition function:
\begin{align}
\hat \rho_h^{\mathcal {D}}(x)=
\sum_{i=1}^n z_i \qquad ,\qquad z_i=e^{-\beta E_i}=
\frac{1}{n h^d} K\left(\frac{x-y_i}{h}\right)
\label{rhD_def2}
\end{align}
where the second equation on the right hand side define the random energies $\beta E_i$. Similar problems have been studied recently in the framework of dense associative memory \cite{lucibello2024exponential} and of generative diffusion \cite{biroli2024dynamical}. 

\subsubsection{New statistical regimes and extreme values dominance.}
From the statistics and physics point of view, the most important transition is the one occurring at the value $h=h_G(\alpha)$, which is a glass transition \cite{derrida1981random}. 
An intuitive way of understanding the phenomenon at play in this glass transition is to focus on the weights 
\begin{align}
    w_i=\frac{K\left(\frac{x-y_i}{h}\right)}{\sum_{j=1}^n K\left(\frac{x-y_j}{h}\right)}\,\,, 
\end{align}
and the so-called participation ratios $Y_k=\sum_i^n w_i^k$.
Their statistics allows to understand the difference between two very different situations: (i) many terms contribute to the sum over $i$ in $\rhD(x)$: their number is large, diverging with $n$, but the contribution of each one of them is very small, vanishing with $n$, (ii) only a few terms contribute to the sum over $i$ in $\rhD(x)$:  the contribution of each one is of the same order of the entire sum. The former situation is the one taking place when $h>h_G(\alpha)$.
In this case $Y_k$ vanishes for $k>1$.  The latter is instead what happens in the regime $0<h<h_G(\alpha)$.  In this case one has the universal result \cite{MPSTV1984,grossmezard1984}:
\begin{align}
    \mathbb{E}_{\mathcal{D}}[ Y_k]=\frac{\Gamma(k-m)}{\Gamma(k)\Gamma(1-m)}
\end{align}
where $\Gamma(z)=\int_0^\infty t^{z-1} e^{-t}dt$ is the Gamma function, and $m$ is a real number in $(0,1)$ defined in Theorem (\ref{thm_concentration_pure}) below. 
The expression above shows indeed that in the 'glassy' regime $0<h<h_G(\alpha)$ the sum over $i$ is dominated by a few terms which are of the order of the entire sum, plus a subleading background due to exponentially many terms each one contributing very little. The existence of this background is revealed by the divergence of $\mathbb{E}_{\mathcal{D}} Y_k$ for $k\rightarrow m$.  In this regime the sum 
in $\rhD$ (see (\ref{rhD_def}))
is dominated by the index $i$ giving the largest {\it extreme values} of  $\frac{1}{n h^d} K\left(\frac{x-y_i}{h}\right)$.  For the isotropic monotonously decreasing kernel we are focusing on, the extreme value corresponding to the maximum term $M$ reads: 
\begin{align}
M=\mathrm{Max}_{i=1,\dots,n}\left[\frac{1}{n h^d} K\left(\frac{x-y_i}{h}\right) \right]=\frac{1}{n h^d} K\left(\frac{d_{min}(x)}{h}\right)\,,
    \end{align}
where $d_{min}(x)$ is the minimal distance between $x$ and each one of the $y_i$s. The next relevant terms in $\rhD$ correspond to the data points $y_i$ ordered in increasing distances to $x$.    

In the asymptotic limit we are considering, $n,d\rightarrow \infty$ with $\alpha$ fixed, 
the density scales exponentially with $d$. In consequence, the quantity which has a good asymptotic limit and concentrates is $\frac{1}{d}\log(\rhD(x))$. It is the analogous of a  "free-entropy" in physics and is like a rate function in large deviation theory. It has a very different behaviour in the regimes outlined above, in particular:
\begin{align}
\mathrm{For}\quad h>h_G(\alpha) \quad \frac{1}{d}\log \rhD(x)=\frac{1}{d}\log \mathbb {E}_{\mathcal{D}}[\rhD(x)]
\end{align}
\begin{align}
\mathrm{For} \quad 0<h<h_G(\alpha) \quad \frac{1}{d}\log \rhD(x)=\frac{1}{d}\log \left(\frac{1}{n h^d} K\left(\frac{d_{min}(x)}{h}\right)\right)
\end{align}
In the former regime, many terms of the sum over the dataset contribute and the Kernel-estimated density concentrates.  
In the latter regime, instead, a few terms dominate the sum over $i$, and they are equal to the largest one at exponential leading order, thus leading to the expression above. This is a highly fluctuating regime in which the Kernel-estimated density does not concentrate and is strongly correlated with the dataset.    

We therefore find that in the large dimensional limit there is a regime in which the usual decomposition in bias and variance holds if the bandwidth is large enough. However, for smaller bandwidth this does not happen any more, and density estimation is governed by extreme statistics. These results are in agreement with the numerical results presented above. When applied to a $\rho$ which is isotropic Gaussian of zero mean and variance $1$, studied with a Gaussian kernel,  our results below predict, for $\alpha=.1$,  $h_{CLT}\simeq 1.61$, and $ h_G\simeq 1.37$. Fig \ref{fig:clt} is a case where $h>h_{CLT}$ whereas Fig \ref{fig:rare} is a case where $h<h_G$.  

Note that in all the previous results $x$ has been a passive-bystander: we have specified from the beginning that $x$ is fixed and drawn from $\rho(x)$. Nevertheless, one could wonder whether a random dependence on $x$ persists. For the distributions $\rho$ that we consider here, the answer is negative - this result, known as self-averaging in the physics of disordered systems, is related to a concentration property emerging in the asymptotic limit $n,d \rightarrow \infty$. For some multimodal distributions, one may have to decompose the distribution into sums of weighted measures, so that the concentration will hold in each measure (see the definition below in Eq.(\ref{mixture})).
\subsubsection{Losses, high-dimensional limit and the relevance of the new statistical regimes}
We now want to discuss the consequences of our results on the losses used to optimize the bandwidth, and more generally, to assess the quality of a Kernel Density Estimation. 
 It is often considered that the mean square error, or $L_2$ loss, is a  cost function which is not suitable for the high-dimensional limit because it does not converge to a well-defined quantity when $d\rightarrow \infty$. The situation, however, is worse than this. In fact, the $L_2$ loss gives misleading results for $h<h_{CLT}$, where the second moment is dominated by atypical samples. This is particularly problematic if the optimal bandwidth for the $L_2$ loss is less than $h_{CLT}$. This is a frequent situation. It occurs  for instance with the high-dimensional isotropic Gaussian case considered above (in this case,one can check that the optimal bandwidth  \cite{epanechnikov1969non} is less than $h_{CLT}$). Similar drawbacks also apply to the $L_1$ loss as the first moment of $\rhD(x)$ is dominated by 
 atypical samples for $h<h_G$. 
 
 In the high-dimensional limit it is better to consider losses which focus on typical samples such as the Kullback-Leibler divergence between $\rhD(x)$ and $\rho(x)$. This is the one we consider in the following.
 One of our important results is  that, in the cases we analyze in the paper, the optimal bandwidth for the KL divergence is in the glass phase $h<h_G$, i.e. in the new statistical regime in which CLT does not apply. This shows the relevance of the methods discussed here to assess the quality of Kernel Density Estimation and obtain the optimal bandwidth in high dimensions. 

 Another possibility, more difficult than KL to analyse, would be to directly focus on the probability that  $|\rhD(x)-\rho(x)|$ is less than a certain value $\epsilon$ for typical samples $x$.

\section{General setup.}
 We have a database of $n$ points $y_i \in \mathbb{R}^d$, with $i=1,...,n$, drawn iid from a
probability density probability law $\rho(x)$, regular enough (in a way to be precised later). We are
interested in the large $d$ and $n$ limit, with $\alpha=(\log n)/d$ fixed.
We want to reconstruct an approximation of $\rho(x)$ from the data
$x_i$, using a kernel $K$. The estimator of the pdf
at a given point $x$ is given by (\ref{rhD_def}).

\subsection{Class of kernels.}
\label{def_kernel}
For simplicity we shall restrict in the following to a specific class of kernels, characterized by a single exponent $\gamma\geq 1$: we define the "$\gamma$-kernel" as
\begin{align}
 K_\gamma (x)=\exp\left(d\left[c_\gamma-\frac{1}{2\gamma} \left(\frac{|x|^2}{d}\right)^\gamma\right]\right)
    \label{gammaKerdef} 
\end{align}
where $c_\gamma$ is a normalization constant ensuring that $\int d^d x K_\gamma (x)=1$. In the large $d$ limit it is given by: 
\begin{align}
    c_\gamma=-\frac{1}{2}\left( \log(2\pi)+1-\frac{1}{\gamma}\right)
    \label{cgammadef}
\end{align}

Note that our approach can be straightforwardly extended to rotational invariant  kernels which are well behaved in the large $d$ limit, in the sense that there exists a "rate function" $f$ such that
$
    K(x)=A e^{-d f(|x|^2/d)}
$
with $f$ regular enough and increasing.

\subsection{Class of densities.}
\label{def_dens}
We need to characterize the scaling of $\rho$ at large $d$. First, we assume that $\rho$ is such
that, $\forall i$, the expectation of $x_i^2$ is of order 1 when $d\to
\infty$. This means that the expectation of $|x|^2$ is $O(d)$.
Let us fix a point $x\in\mathbb{R}^d$ chosen randomly from the distribution with density $\rho$, and a value of $h$. We now generate $y$ distributed according to $\rho$, and we consider the random variable $u=|x-y|^2/(d h^2)$, which is of order $O(1)$ when $d\to \infty$. It has a probability density which we call $f_{d,x,h}(u)$. Let us define its cumulant-generating function as
\begin{align}
    \tilde f_{d,x,h}(\lambda)=\frac{1}{d}\log \left[\mathbb{E}_u \; e^{-d \lambda u}\right]\ .
\end{align}
\begin{definition}[Pure densities]
This generating function $\tilde f_{d,x,h}(\lambda) $ is a random function which depends on $x$. We will define a 'pure' density $\rho$ as a density such that this generating function concentrates.  More precisely, for pure densities, the distribution of $\tilde f_{d,x,h}(\lambda) $, when $x$ is sampled from $\rho$, concentrates around its mean $\overline{f}_{d,h}(\lambda)=\int dx \rho(x)\tilde f_{d,x,h}(\lambda)$  in the limit $d\to \infty$.
\end{definition}

Accordingly, 
for pure densities with well-behaved $\overline {f}$, given a bandwidth $h$ and a typical $x$ generated from $\rho$, the distribution of $u=|x-y|^2/(d h^2)$
 satisfies a large deviation principle $P(u)\sim e^{-d J_h(u)}$ with a rate function $J_h(u)$ given by the Legendre transform and its inverse :
 \begin{align}
     \overline{f}_{d,h}(\lambda)= -\min_u[\lambda u+ J_h(u)] \ \ \text{and}\ \ \ J_h(u)=-\min_\lambda[\lambda u+\overline{f}_{d,h}(\lambda) ]
     \label{eq:rate_func_u}
 \end{align}

There are many examples of pure densities; for instance multivariate Gaussians (with a covariance matrix having a well defined limit of the density of eigenvalues at large $d$), or densities with independent components, or  statistical-physics inspired models where the variables $x_i$, $x_j$ interact whenever the two points $i,j$ are neighbours on a $D$-dimensional grid, considered in their high temperature phases (e.g. Ising models).

Our approach can be extended to probabilities which are mixtures of pure densities. These are densities which can be written as 
\begin{align}
    \rho=\sum_{r=1}^k w_r \rho_r
    \label{mixture}
\end{align}
 where $w_r$ are positive weights normalized to $\sum_{r=1}^k w_r=1$, and the $\rho_r$ are pure densities which satisfy some cross-concentration properties. We shall restrict to mixtures where $k$ is finite when $d\to\infty$. When $x$ is sampled from $\rho_r$, we denote by $\tilde \rho^r_{d,x,h}(u)$ the probability density of $u=|x-y|^2/(d h^2)$. As $y$ is sampled from $\rho=\sum_s w_s \rho_s$, this probability density can be written as 
\begin{align}
  \tilde\rho^r_{d,x,h}(u) = \sum_{s=1}^k w_s F^{rs}_{d,x,h}(u)
\end{align}
and we can introduce the cumulant-generating functions
\begin{align}
    \tilde F^{rs}_{d,x,h}(\lambda)=\frac{1}{d}\log \left[\mathbb{E}_{u\sim  \rho_s} \; e^{-d \lambda u}\right]
\end{align}
The fact that $\rho_r$ is a pure density means that $F^{rr}$ concentrates. The mixed densities generalize this statement to the following case:
\begin{definition}[Mixed densities]
    A mixture of pure densities is a density which can be decomposed as a sum of pure densities $\rho=\sum_{r=1}^k w_r \rho_r$
    such that, for all $r,s\in\{1,...,k\}^2$ the distribution of $\tilde F^{rs}_{d,x,h}(\lambda) $, when $x$ is sampled from $\rho_r$, concentrates around its mean $\int dx \rho_r(x)\tilde F^{rs}_{d,x,h}(\lambda)$ in the limit $d\to \infty$.
\end{definition}
In the following we shall focus our study on the case of pure densities, and only briefly mention an example of generalization to mixed densities.

\subsection{Definition: "replica free entropy".}
Let us introduce the function
\begin{align}
    g(x,h^2,m)=\frac{1}{d} \log \left\{ \int \frac{d^d y}{h^{md}} \; \rho(y) \left[K\left(\frac{x-y}{h}\right)\right]^m\right\}
  =\frac{1}{d}\log \mathbb{E}_y K^m\left(\frac{x-y}{h}\right) -m\log h
    \label{free_ent_def}
\end{align}
where $\mathbb{E}_x$ means the expectation value with $x$ sampled from the probability density function$\rho(x)$.

When $\rho$ is a pure density, $x$ is sampled according to  $\rho$, and $K$ is a $\gamma$-kernel, we shall see that for all $h>0$ and for all 
$m>0$, the distribution of this random variable concentrates at large $d$ around its mean, which has a well defined limit 
\begin{align}
    \overline{g}(h^2,m)=\lim_{d\to\infty}\frac{1}{d}\mathbb{E}_x\log \mathbb{E}_y K^m \left(\frac{x-y}{h}\right)-m\log h
    \label{gbarh2m}
\end{align}
Note that both $g(x,h^2,m)$ and $\overline{g}(h^2,m)$ are convex functions of $m$, and we shall also assume that the density $\rho$ is such that $\overline{g}(h^2,m)$ is strictly convex.

\begin{definition}
We define the 'replica free entropy' $\phi_{\alpha,h}(m)$ as
\begin{align}
    \phi_{\alpha,h}(m)= \frac{1-m}{m}\alpha +\frac{1}{m} \overline{g}(h^2,m)\ .
\end{align} 
\label{def_repF}
\end{definition}

Notice that $\phi_{\alpha,h}(1)=\frac{1}{d}{\mathbb{E}}_{x}\log\left\{ {\mathbb{E}}_{\mathcal{D}}\hat \rho_h^\mathcal{D}(x)\right\}$ gives the leading exponential behavior of the average kernel estimate of $\rho$ at a generic value of $x$.

\section{Results}
\subsection{Central limit theorem transition.}

The following result  asserts the existence of a critical value of the bandwidth above which the standard deviation of $\rhD(x)$ is much smaller than the typical scale of $\rhD(x)$.
\begin{proposition}
    \label{propo_CLT}
 In the large dimensional limit, 
 when $\rho$ is a pure density and $K$ is a $\gamma$-kernel,  there exists a critical value of the bandwidth, $h_{CLT}(\alpha)$ which is the unique solution of
 \begin{align}
 \overline g(h^2,m=2)-2\overline g(h^2,m=1)=\alpha
 \label{eq:hCLT}
 \end{align}
 This critical value of the bandwidth is such that,  when $x$ is a point sampled from the density $\rho$:
 \begin{align}
 \text{If}& \ h>h_{CLT}(\alpha)\ \ : \ \ \lim_{d\to\infty}\frac{1}{d} \log  \frac{ \text{var} \; \rhD(x) }{[\mathbb{E}_{\mathcal{D}} \; \rhD(x)]^2} \; <\; 0\nonumber\\
 \text{If}& \ h<h_{CLT}(\alpha)\ \ : \ \ \lim_{d\to\infty}\frac{1}{d} \log  \frac{ \text{var} \; \rhD(x) }{[\mathbb{E}_{\mathcal{D}} \; \rhD(x)]^2}\; >\; 0
 \end{align}
\end{proposition}
\subsection{Glass transition}
Let us introduce the
 derivative of the replica free entropy at $m=1$:
 \begin{align}
          D(\alpha,h)&=\frac{\partial \phi_{\alpha,h}(m)}{\partial m}\;(m=1)
   =-\alpha-\mathbb{E}_x\frac{1}{d}\log \mathbb{E}_y K\left(\frac{x-y}{h}\right)+
   \mathbb{E}_x\frac{1}{d}\frac{
   \mathbb{E}_y K\left(\frac{x-y}{h}\right)\log K\left(\frac{x-y}{h}\right)
   }
     { \mathbb{E}_y K\left(\frac{x-y}{h}\right)}
\end{align}
Then we have the following results:
\begin{lemma}
    \label{lemma_hc}
Using a $\gamma$-kernel, if $\rho$ is a pure density,
the equation $D_{\alpha,h}=0$ defines, in the $\alpha,h$ plane, a critical line $h_G(\alpha)$ such that
$D_{\alpha,h}<0$ when $h>h_G(\alpha)$ and $D_{\alpha,h}>0$ when $h<h_G(\alpha)$. The critical value $h_G(\alpha)$, defined by
\begin{align}
    D_{\alpha,h_G(\alpha)}=0
\end{align}
is an increasing function of $\alpha$.
\end{lemma}

We can get a more explicit expression for $h_G(\alpha)$, using the rate function $J_1(u)$ defined in (\ref{eq:rate_func_u}) which
describes the distribution over $y$ of $u=|x-y|^2/d$ when $x$ is a given typical point sampled from $\rho$. Then one has $\overline g(h^2,m=1)=\max_u\left[ c_\gamma -J_1(u)-(1/(2\gamma)) (u.h^2)^\gamma\right]$. When $J_1$ function is well behaved (eg when it is convex) this function has a single maximum at  $u=u^*(h)$, and $ D_{\alpha,h}=J_1(u^*(h))-\alpha$. Then $h_G$ is found by solving $J_1(u^*(h_G))=\alpha$.

\begin{theorem}
\label{thm_concentration_pure}
When $x$ is sampled according to a pure density $\rho$, the distribution of $(1/d)\log \hat \rho_h(x)$ obtained using a $\gamma$-kernel with $\gamma\geq 1$ concentrates at large $d$ around its mean $f$, which is equal to:
\begin{align}
   f&=\phi_{\alpha,h}(m=1)= \frac{1}{d}\log\left({\mathbb {E}}_{\mathcal{D}} \rhD(x)\right) \ \ \ \text{if}\ \ \ h>h_G(\alpha)\label{free_large_h}\\
  f
&=\phi_{\alpha,h}(m=m^*)<\frac{1}{d}\log\left({\mathbb {E}}_{\mathcal{D}} \rhD(x)\right)  \ \ \ \text{if}\ \ \ h<h_G(\alpha) \label{free_small_h}
\end{align}
where $m^* $ is given by the unique solution of $d\phi_{\alpha,h}/dm=0
$ in the interval $(0,1)$.
\end{theorem}

Using the language of statistical physics,
we shall call the phase where $h>h_G(\alpha)$ a ``replica symmetric'' (RS)
phase. This is the phase where  the empirical density
$\hat\rho_h^{\mathcal{D}}$ concentrates at large $d$ around
its expectation value, to exponential accuracy:
\begin{align}
  \lim_{d\to\infty}(1/d)\log \hat\rho_h^{\mathcal{D}}=
  \lim_{d\to\infty}(1/d)\log \left[
  \mathbb{E}\hat\rho_h^{\mathcal{D}}
  \right]
\end{align}
In statistical physics terminology, this identity is referred to as the  equality of the
quenched and annealed averages.
The phase where  $h<h_G(\alpha)$ is called a
`` one-step replica symmetry breaking'' (1RSB)
phase. In this  phase the logarithm of the empirical density
$(1/d)\log\hat\rho_h^{\mathcal{D}}$ concentrates around a value 
which is different from the  logarithm of the expectation value of
$\rho$: the fluctuations of $\hat\rho_h^{\mathcal{D}} $ become too large
and the first moment estimate is not accurate.

\begin{corollary}
Under the hypotheses of Theorem \ref{thm_concentration_pure}, the Kullback-Leibler divergence 
\[
D_{KL}(\rho||\hat \rho_h)=\int dx \rho(x) \log \frac{\rho(x)}{\hat \rho_h(x)}
\]
is given by:
\begin{align}
    D_{KL}(\rho||\hat \rho_h)&=\int dx \rho(x) \log \rho(x)-\phi_{\alpha,h}(m=1)   \ \ \ \text{if}\ \ \ D(\alpha,h)<0 \label{DKL_RS}\\
    D_{KL}(\rho||\hat \rho_h)&=\int dx \rho(x) \log \rho(x)-\phi_{\alpha,h}(m=m^*)   \ \ \ \text{if}\ \ \ D(\alpha,h)>0 \label{DKL_1RSB}
\end{align}
Furthermore,
if the rate function $J_1(u)$ verifies $2\frac{dJ_1(u)}{du}u>-1$ for $u_{G}<u<u_{typ}$, where $u_G$ is the unique solution of $J_1(u_G)=-\alpha$ and $u_{typ}$ is the unique solution of $\frac{dJ_1(u_{typ})}{du}=0$, then the optimal value $h_{opt}$ of the bandwidth, which corresponds to the minimum with respect to $h$ of $D_{KL}$, is reached in the 1RSB regime $D(\alpha,h)>0 $ and satisfies the equation:
\begin{align}
\frac{\partial \phi_{\alpha,h_{opt}}(m^*)}{\partial h}=0  
\end{align}
\label{coroll_DKL_pure}
\end{corollary}

 As we shall show below, this is indeed the case for multi-variate high-dimensional Gaussian distributions. 

\subsection{Probability distribution of $\rho_h^{\mathcal{D}}$.}
\label{sec:proba_distrib_Z}
The previous results state the existence of three regimes when varying $h$, separated by two characteristic values of $h$: $h_{\rm{CLT}}$ and $h_{G}$. The probability distribution of $\rho_h^{\mathcal{D}}$ is very different in these three regimes. Note that $\rho_h^{\mathcal{D}}$ is (at fixed $x$) a sum of independent random variables, so one could expect concentration towards universal $\alpha$-stable laws. This is indeed what happens but the phenomenon is a subtle one as the distribution of the random variables scales with their number,  which makes the framework quite different from the usual one.  We find the following results. 
\begin{corollary}
Under the hypotheses of Theorem (\ref{thm_concentration_pure}) and for $h>h_{CLT}(\alpha)$ the distribution of 
\begin{align}
g=\frac{\rhD(x)-{\mathbb {E}}_{\mathcal{D}}[\rhD(x)]}{\sqrt{\mathrm{Var}_{\mathcal{D}}[\rhD(x) ]}}
\end{align}
converges in law to a Gaussian distribution with unit variance and mean zero. 
\label{thm_gauss}
\end{corollary}

This result extends the standard CLT regime which holds for Kernel Density Estimation when $d$ is fixed and $n\rightarrow \infty$,  to the high-dimensional case when the bandwidth is large enough. 
Instead, for $h<h_{CLT}(\alpha)$, this does not happen any longer: the centred and rescaled $\rhD(x)$ converges instead to a $\overline{\alpha}$-stable with $\overline{\alpha}<2$. 
\begin{corollary}
Under the hypotheses of Theorem (\ref{thm_concentration_pure}),  for $h<h_{CLT}(\alpha)$ the distribution of 
\begin{align}
l&=\frac{\rhD(x)-
{\mathbb {E}}_{\mathcal{D}}[\rhD(x)]
}{\exp({\mathbb {E}}_{\mathcal{D}}[\log \hat\rho_{h,2}^{\mathcal{D}}(x)])}\qquad \mathrm{for}\,\, h_G(\alpha)<h<h_{CLT}(\alpha)\\
l&=\frac{\rhD(x)
}{\exp({\mathbb {E}}_{\mathcal{D}}[\log \hat\rho_{h,1}^{\mathcal{D}}(x)])}\qquad \mathrm{for}\,\, h<h_G(\alpha)
\end{align}
converges in law to a $\overline{\alpha}$-stable distribution with skewness $\beta=1$ and $\overline{\alpha}=m^*$, where $m^* $ is given by the unique solution of $d\phi_{\alpha,h}(m)/dm=0
$ in the interval $(0,2)$ and $\hat\rho_{h,p}^{\mathcal{D}}=\left(\sum_{i=1}^nz_i^p\right)^{1/p}$ where $z_i$ is defined in eq. (\ref{rhD_def2}).
\label{thm_alpha}
\end{corollary}
The most important feature of the distribution of $l$ is its power-law behavior at large $l$: $P(l)\sim 1/l^{1+m^*}$. 
In the regime $h<h_{G}(\alpha)$, $m^*<1$ and the
distribution of $l$ does not have a  first moment. The average value of $\rhD(x)$ exists but it is different from its typical value. This phenomenon, and more generally the results quoted above, can be understood following Ref. \cite{gardner1989probability}. The distribution of $l$ has two parts: one describing fluctuation on the scale $l\sim O(1)$ and another one capturing fluctuations of $(\log l)/d \sim O(1)$. The former is the $\overline{\alpha}$-stable law discussed above. The latter has a large deviation form: $\exp(d f((\log l)/d ))$. Depending on the kind of average and the regime of $h$ one focuses on, it is the former or the latter that gives the dominant contribution. For $h<h_G(\alpha)$, the typical value of $l$ is determined by the former, but the average by the latter. For  $h_G(\alpha)<h<h_{CLT}(\alpha)$ the typical value of $l$ is determined by the former, but the variance by the latter (the average is zero). Figure \ref{fig:clt} and \ref{fig:rare} show a concrete numerical example of the results stated in this section. \\

The proof of these results can be obtained by a generalization of the approach in \cite{bovier2002fluctuations,ben2005limit}. In the physics literature on spin-glasses, the corresponding results have been obtained in the 80s, see \cite{mezard1987spin}. The main proofs related to this section will be given in Sect.\ref{proofthmain}.

\section{A detailed example: Gaussian density}\label{sec:gaussexa}
Let us study the case where $\rho$ is a centered Gaussian, with a covariance matrix $C$ which has a density of eigenvalues that goes to a well defined limit in the large $d$ limit. This means that, if we call $c_r$ the $r$-th eigenvalue, then the density $(1/d) \sum_r \delta(\lambda-c_r)$ goes to a well defined limit $\rho_C(\lambda)$, in the sense of distributions. The resulting $\overline{g}(h^2,m)$ can be computed for a general $\gamma$-kernel.

\subsection{Results}
We state here the results for general distribution of eigenvalues $\rho_C(\lambda)$ and $\gamma$-kernel. 
The proofs are given in Sect.\ref{GaussianDensity_proofs}.

\subsubsection{Concentration}
\begin{proposition}
\label{prop_Concentration_Gauss}
The Gaussian density $\rho(x) $ is a pure density. The replica free entropy computed with a $\gamma$-kernel is:
\begin{align}
\phi_{\alpha,h}(m)=&\frac{1-m}{m}\alpha+ c_\gamma-f_\gamma\left(\frac{l}{h^2}\right)- \log h\nonumber\\ 
&-\frac{1}{2m}\int d\lambda \rho_C(\lambda)\log(1+\hat l \lambda)
-\frac{1}{2m}\int d\lambda \rho_C(\lambda)\frac{\hat l\lambda}{1+\hat l\lambda}
+\frac{\hat l l}{2m}\ ,
\label{phi_Gauss}
\end{align}
where $f_\gamma(u)=u^{\gamma}/(2\gamma)$, and
 the  two variables $l$ and $\hat l$ are the unique solution of the two equations expressing the stationarity of the replica free entropy $\phi_{\alpha,h}(m)$ with respect to $l,\hat l$:
\begin{align}
    \frac{\hat l}{2}=\frac{m}{h^2}f_\gamma'\left(\frac{l}{h^2}\right)\,\,\,\,,\,\,\,\,
    l=\int d\lambda \rho_C(\lambda) \left(\frac{\lambda}{1+\hat l \lambda}+\frac{\lambda}{(1+\hat l \lambda)^2}\right)\ .
    \label{lhatl}
\end{align}
    \end{proposition}

These equations are easily solved numerically for arbitrary $\gamma$-kernels. They can be solved explicitly when the kernel is Gaussian ($\gamma=1$). One then finds $\hat l=m/h^2$, and $l$ is expressed as a function of the bandwidth $h^2$ and the eigenvalue density $\rho_C$. This gives:
\begin{align}
\phi_{\alpha,h}(m)&=\frac{1-m}{m}\alpha+\frac{1-m}{m}\log h  -\frac{1}{2}\log (2\pi)
\nonumber
\\
&-\frac{1}{2m}\int d\lambda \rho_C(\lambda)\log(h^2+m \lambda)
-\frac{1}{2}\int d\lambda \rho_C(\lambda)\frac{\lambda}{h^2+m\lambda}
\label{phi_Gauss_kernel}
\end{align}

    \subsubsection{Phase diagram}
\begin{proposition}
\label{prop_CL_Gauss}
 For general $\gamma$-kernels, the critical line for the glass transition, $h_G(\alpha)$ is defined by the unique solution of $D(\alpha,h_G(\alpha))=0$ where the function $D(\alpha,h)$ is given by
\begin{align}
    D(\alpha,h)=-\alpha +\frac{1}{2}\int d\lambda \; \rho_C(\lambda)\; \left(
  \log(1+\hat l^* \lambda)- \frac{\hat l^*\lambda}{(1+\hat l^*\lambda)^2}  \right)\ . 
  \label{D_Gauss}
\end{align}
where $\hat l^*$ is the value of  $\hat l$ which solves equations (\ref{lhatl}) with $m=1$.
\end{proposition}

For Gaussian kernels we have $\hat l^*=1/h^2$ and the equation giving the bandwidth of the glass transition  $h_G(\alpha)$ simplifies to 
\begin{align}
    \alpha =+\frac{1}{2}\int d\lambda \; \rho_C(\lambda)\; \left(
  \log(1+\lambda/h^2)- \frac{\lambda/h^2}{(1+ \lambda/h^2)^2}  \right)\ . 
  \label{D_Gauss_Gauss}
\end{align}
Fig. \ref{fig:gauss_comp} plots $h_{CLT}(\alpha)$ (obtained from \ref{eq:hCLT}) and $h_G(\alpha)$ obtained from (\ref{D_Gauss_Gauss}) in the case where the covariance is the identity matrix. In this case, one can check that, in the large $\alpha$ limit one has $h_{CLT}\simeq \sqrt{e/2} e^{-\alpha}$ and $h_{G}\simeq  e^{-\alpha}$.

We can now compute the KL divergence. For a given $\alpha$, calling $\hat l
$ the solution of 
 \begin{align}
   \alpha&=\frac{1}{2}\int d\lambda \rho_C(\lambda) \left(
\log(1+\lambda\hat l)-\frac{\lambda\hat l}{(1+\lambda\hat l)^2}
   \right)
   \end{align}
and $l$ given by 
 \begin{align}
     l=\int d\lambda \rho_C(\lambda ) \left(\frac{\lambda}{1+\hat
          l \lambda}+\frac{\lambda}{(1+\hat l
          \lambda)^2}\right)\ ,
  \end{align}
 we have:
\begin{proposition}
\label{prop_KL_Gauss}
When $h<h_G(\alpha)$, the KL divergence $D_{KL}(\rho||\hat\rho_h^{\mathcal{D}})$  is 
equal to:
\begin{align}
 \lim_{d\to\infty} \frac{1}{d} D_{KL}(\rho||\hat \rho_h)=\alpha-\frac{1}{2d}\langle {\text{Tr}} \log C\rangle +\frac{1}{2}\log u^* -f(u^*)+ \log h+f\left(\frac{l}{h^2}\right)
 \label{KLgauss}
\end{align}
\end{proposition}
\subsubsection{Optimal bandwidth}
\begin{proposition}
\label{prop_KL_opt_Gauss}
The KL divergence $D_{KL}(\rho||\hat\rho_h^{\mathcal{D}})$  given in
Corollary \ref{coroll_DKL_pure} is a function of $h$ which has a
minimum at a value $h^*(\alpha)$ which is the solution of 
\begin{align}\label{eq:hopt}
    \frac{l}{h^2}f'\left( \frac{l}{h^2}\right)=\frac{1}{2} \ .
\end{align} 
The optimal width of the kernel $h^*(\alpha)$ is smaller than $<h_G(\alpha)$. Therefore the optimal
bandwidth is obtained in the 1RSB phase. The optimal KL divergence obtained by choosing  $h=h^*(\alpha)$ is equal to 
\begin{align}\label{eq:DKLopt}
 \lim_{d\to\infty}  \frac{1}{d} D_{KL}^{min}= \alpha-\frac{1}{2} \langle \log \lambda \rangle+
 \frac{1}{2}\log l
\end{align}
 \end{proposition}

We notice that the minimal $D_{KL}$ is independent of the kernel. This result can be understood  using the fact that, in the RSB phase:
\begin{align}
\frac{1}{d}\rho_h(x)=\exp\left(-d\left(-c_\gamma+\alpha+\log h +f(d_{min}^2/h^2) \right)\right)
\end{align}
where $d_{min}$ is the minimum (intensive) distance between the point $x$ (drawn at random from the distribution $\rho$) and the $n$ points $y_\mu$ (also drawn at random from the distribution $\rho$). \\
Since the only $h$-dependent part of $D_{KL}$ is given by $\frac{1}{d}\rho_h(x)$, one has to optimize this term with respect to $h$, which leads to the equation: 
\begin{equation}\label{eq:ustar}
2f'\left(\frac{d^2_{min}}{h^2} \right)\frac{d^2_{min}}{h^2}=1
\end{equation}

Note that this is nothing else than eq. (\ref{eq:hopt}), and hence provides an interpretation for $l$ in that equation.  \\
Moreover, using the normalization equation of the kernel dependent contribution to $D_{KL}$ one finds:
 \[
\frac{1}{d}\rho_h(x)=\exp\left(-d\left(\alpha+\frac{1}{2}\log d_{min}^2-\frac{1}{2}\log (d_{min}^2/h^2) +f(d_{min}^2/h^2) +\frac{1}{2\gamma }+\frac{1}{2}\log(2\pi e)\right)\right)
\]
By noticing that $u^*$ satisfies the same equation than $d_{min}^2/h^2$ (see eq.  \ref{eq:ustar}),  one finds that the third and fourth terms cancel with the fifth and the sixth ones, thus leaving a kernel-independent contribution.  In summary, at the optimal bandwidth, we find that 
\[
\frac{1}{d}D_{KL}=-\frac{1}{d}S(\rho)+\alpha+\frac{1}{2}\log(2\pi e)+\frac{1}{2}\log d_{min}^2
\]
where $S(\rho)$ is the Shannon entropy of the distribution $\rho$.  By replacing the entropy of a multivariate Gaussian, one indeed finds back eq.  (\ref{eq:DKLopt}).\\
The value of the optimal bandwidth depends on the kernel, and is equal to 
\[
h^2_{opt}=d_{min}^2/x^2_{typ}
\]
where $x^2_{typ}=u^*$ is the typical distance of a point drawn from a single kernel term $K(x)$.

\subsubsection{Numerical test}
\begin{figure}
    \centering
    \includegraphics[width=0.38\textwidth]{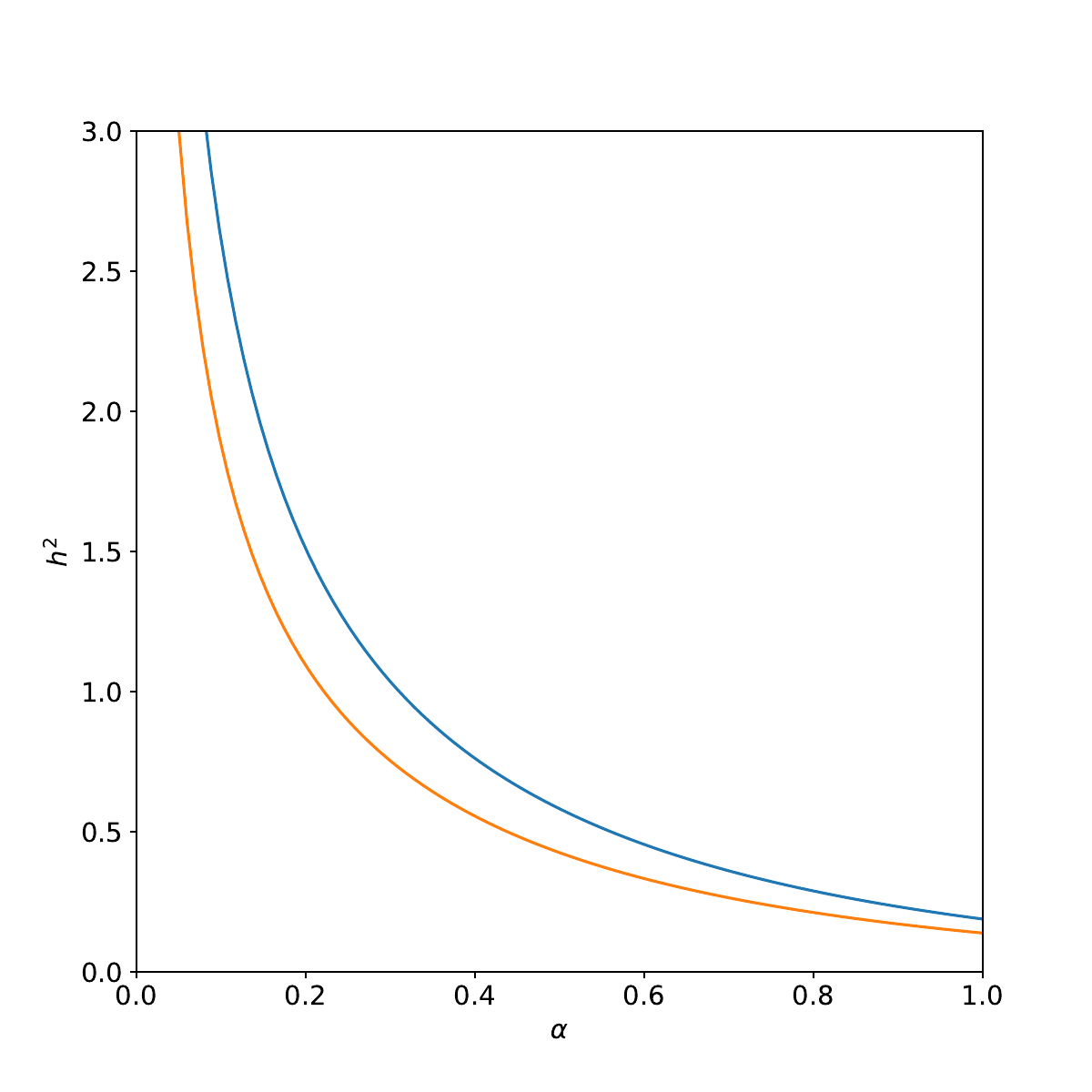}
    \includegraphics[width=0.58\textwidth]{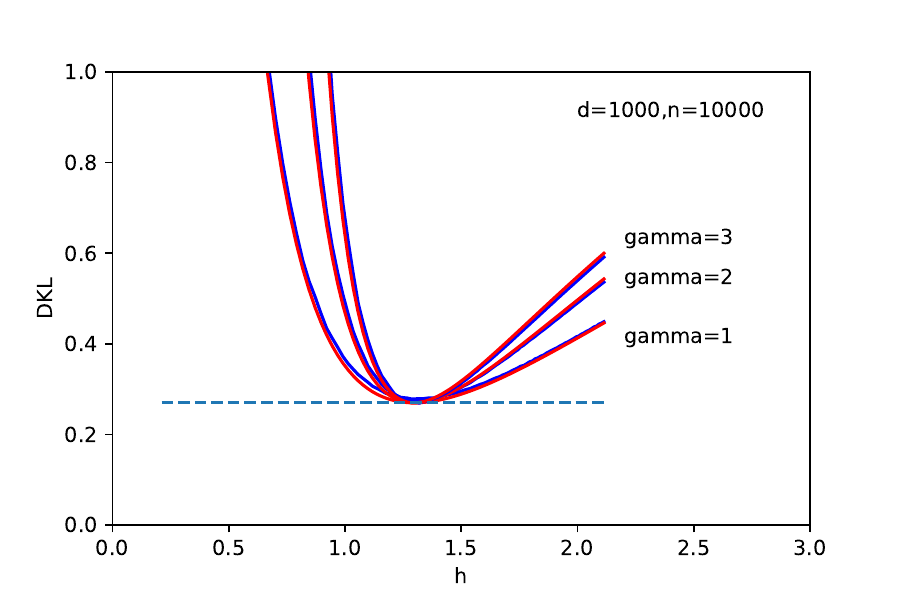}
    \caption{The case where $\rho$ is a Gaussian with identity covariance matrix. Left: $h_{CLT}^2$ (top curve) and $h_G^2$ are plotted as function of $\alpha$ for a kernel with $\gamma=1$. Right: For each of the three kernels with $\gamma=1,2,3$, plot of $D_{KL}$ versus $h$, comparison of the theory (red curves) and the experiment (blue curves) described in the text.}
    \label{fig:gauss_comp}
\end{figure}

One can test numerically the prediction for $D_{KL}$ contained in Propositions \ref{prop_KL_Gauss},\ref{prop_KL_opt_Gauss}. Taking a Gaussian $\rho(x)$ in dimension $d=1000$ with a covariance matrix equals to identity, we have generated $n=10,000$ random points.  This database is used in order to define $\hat \rho_h(x)$. Then, in order to estimate $-\int dx \rho(x) \log\hat \rho_h(x) $, one generates $M$ new points $y_j$, $j=1,...,M$ sampled iid from $\rho$ and one computes $(1/M)\sum_{j=1}^M \log\hat \rho_h(y_j)$. The simulation is done with $d=1000$, $n=10,000$ and $M=200$. Fig.\ref{fig:gauss_comp} shows a comparison between the KL divergence determined numerically and the analytical prediction of (\ref{KLgauss}). The comparison is done with kernels defined by $f(x)=x^\gamma/(2 \gamma)$, with $\gamma=1,2,3$. Note the very good agreement between theory and numerics. Our theory works well also in a regime in which $d\gg \log n$, i.e. the regime of small $\alpha$s (the corresponding value of $\alpha$ for this numerical example is $\alpha=0.004$).

\section{Proofs}
\label{proofthmain}
\subsection{Proof of Proposition \ref{propo_CLT}}
We shall compute the first two moments of $\rhD$.
Under the hypotheses of Sect. \ref{def_dens}, when $x$ is a point sampled from the density $\rho$, the expectation value over $\mathcal{D}$ of the estimator $\rhD(x)$ is given, to leading exponential order at large $d$, by
\begin{align}
 \mathbb{E}_{\mathcal{D}} \; [\rhD(x)] \simeq e^{d \overline g(h^2,m=1)}
\end{align}
(Here and in the following we denote the equality to leading exponential order by $\simeq$.  $A(d)\simeq B(d)$ means that $\lim_{d\to\infty}\frac{1}{d} \log A$  $=$  $ \lim_{d\to\infty}\frac{1}{d} \log B $).

We now compute the variance (over the choice of the data) of $\rhD(x)$, for a typical $x$, drawn from the density $\rho$. Expressing $\rhD(x)^2$ as a double sum over two datapoints $i,j$ in $\mathcal{D}$, and distinguishing the case $i=j$ and $i\neq j$, we get:
\begin{align}
 \text{var} \; \rhD(x) \simeq e^{-d \alpha}\left[ e^{d \overline g(h^2,m=2)}-
 e^{2 d \overline g(h^2,m=1)}\right]\ .
\end{align}
Therefore:
\begin{align}
\lim_{d\to\infty}\frac{1}{d} \log  \frac{ \text{var} \; \rhD(x) }{[\mathbb{E}_{\mathcal{D}} \rhD(x)]^2}=\overline g(h^2,m=2)-2 \overline g(h^2,m=1)-\alpha\ .
 \end{align}

We use the fact that $\overline g(h^2,m)=m c_\gamma -m \log h+G(m/h^{2\gamma})$,
with
\begin{align}
\label{Gdef}
    G(u)=\frac{1}{d}\; \mathbb{E}_x\; \log \left[\mathbb{E}_y \; \exp\left(-u\frac{d}{2\gamma}[(x-y)^2/d]^\gamma\right)\right]\ .
\end{align}
The function $G(u)$ is convex and monotonously decreasing on $u\in(0,\infty)$. It satisfies $G(0)=0$, $G(u)\to -\infty$ when $u\to \infty$.
Therefore the function $u\to G(2u)-2 G(u)$ is monotonously increasing from $0$ to $\infty$ when $u$ goes from $0$ to $\infty$, and the equation $\overline g(h^2,m=2)-2 \overline g(h^2,m=1)-\alpha=0$ has a unique solution $h_{CLT}(\alpha)$.  $\square$

\subsection{Proof of the theorems concerning the glass transition } 

\subsubsection{Summary of the Random Energy Model}
\label{app:rem1}
The Random energy model is a simple model of disordered system which was introduced and originally studied by Derrida\cite{derrida1981random}. Here we briefly summarize some of the main known results of the REM, in a generalized case studied in \cite{BouchaudMezard97}. This  presentation is partially based on \cite{lucibello2023exponential}, with an adaptation of notations to the present case. More formal proofs of the results can be found in \cite{olivieri1984existence,ben2005limit}. For a more extended introduction, the reader can also consult chapters 5 and 8 of \cite{mezard2009information}.

Consider a set $\cS$ of $n=e^{\alpha d}$ independent random variables $\varepsilon^\mu$
which are i.i.d.  random variables with probability density function (pdf) $p_d(\varepsilon)$.  This pdf is assumed to satisfy, at large $d$, a large deviation
principle with a rate function $I(\ve)$.
That is, for any $a<b$:
\begin{align}
\lim_{d\infty} \frac{1}{d} \log \int_a^b d\ve \; p_d(\varepsilon)\approx  -\inf_{\ve \in [a,b]} I(\ve).
\label{eq:rate-s}
\end{align}
A classical choice  \cite{derrida1981random} for the pdf is $p_d = \mathcal{N}(0, 1/d)$, which results in $I(\ve)=\ve^2$. We shall use a generalized form, but keep for simplicity  to cases where the function $I(\ve)$ is a
strictly convex non-negative function, reaching $0$ at one single value $\ve_0$.

The central object of study in the REM is the so-called partition function defined as 
\begin{align}
    Z_\cS=\sum_{i=1}^n e^{- d \ve_i}
    \label{Zremdef}
\end{align}
In physics, the independent random variables $
\ve_i $are called energies, hence the name of the model. In Boltzmann's formalism (here at inverse temperature $1$), one defines the probability that the system occupies the energy level $\ve_i$ as $p_i=\frac{1}{Z_\cS} e^{-d \ve_i}$, and the computation of the partition function is an important step in the understanding of the properties of this probability distribution. 

One can define the free-entropy  of the REM as $\Phi_\cS=\frac{1}{ d}\log Z_\cS$.
A main consequence of the independence of the variables in $\cS$, and of the specific
large-deviation form of their distribution is that, in the large $d$
limit, the random variable $\Phi_\cS$ concentrates: its distribution
becomes peaked around its typical value\cite{olivieri1984existence}
\begin{equation}
    \phi= \lim_{d\to\infty} \mathbb{E}\,\Phi_\cS\ .
\end{equation}
where $\mathbb{E} $ denotes the expectation with respect to the choice of the database $\cS$.
 Let us see how one can
compute the typical value of the
free energy density in the large $d$ limit, $\phi$,
which depends on $\alpha,\lambda$ and on the rate function $I(\ve)$, and justify the concentration property.
In the large $d$ limit, let us call $\mathcal{N}_{[a,b]}$ the number of random variables of $\cS$ (among the $n=e^{\alpha d}$ it contains) which
are in the interval $ [a,b] $. To leading exponential order, its expected value is 
\begin{align}
\mathbb{E}\, \mathcal{N}_{[a,b]}\simeq e^{d(\alpha - \min_{\ve \in [a,b]}
  I(\ve))},
\end{align}
therefore the average density of random variables around $\ve$ is $e^{d(\alpha -  I(\ve))}  $. 
The function $\alpha -  I(\ve)$ is a concave
function of $\ve$, it vanishes at two values $\ve=\ve_0$ and $\ve=\ve_1$, with
$\ve_0<\ve_1$ (obviously $\ve_0$ and $\ve_1$ depend on $\alpha$, we do not write this dependence explicitly  in order to lighten the notations), it is positive for $\ve \in [\ve_0,\,\ve_1]$ and it is
negative outside of this interval. 
Using the first- and second-moment methods, one can prove that at large
$d$, $\frac{1}{d}\log 
\mathcal{N}_{[a,b]}$ concentrates around $\max_{\ve \in
  [a,b]}\psi_\alpha(\ve)$, where:
\begin{equation}
\psi_\alpha(\ve) =
\begin{cases}
\alpha -  I(\ve) & \text{if } \ve \in [\ve_0,\,\ve_1]\\
-\infty  & \text{otherwise}
\end{cases}
\label{eq:psityp}
\end{equation}
We shall not detail this proof, referring the reader to \cite{derrida1981random}. Let us just mention the ideas behind
the proofs. The first
moment method uses Jensen's inequality $ \log 
\mathbb{E}\mathcal{N}_{[a,b]}  \geq \mathbb{E} \log 
\mathcal{N}_{[a,b]}$ in order to show that, when
$\alpha -  I(\ve) <0$, the average number of variables in $\cS$ around
$\ve$ is exponentially small in $N$, which implies that their typical
number  is zero. This explains the $-\infty$ case in Eq. \eqref{eq:psityp}. The second
moment method uses the independence of the variables to show that,
when $ \mathbb{E} \mathcal{N}_{[a,b]}$ is exponentially large in $N$, the relative fluctuations
$ \sqrt{\mathbb{E} (\mathcal{N}_{[a,b]}^2)-(\mathbb{E} \mathcal{N}_{[a,b]} )^2}/ (\mathbb{E} \mathcal{N}_{[a,b]})$
are exponentially small, leading to the concentration result \eqref{eq:psityp}.

Let us now study the partition function \eqref{Zremdef}. Using the
concentration property for the density of levels
\eqref{eq:psityp}, one obtains the large $d$ behaviour
\begin{align}
Z\approx\int_{\ve_0}^{\ve_1}d\ve \; e^{d (\alpha
  -I(\ve)- \ve)}.
  \label{eq:REM_Zint}
  \end{align}
This integral can be evaluated using Laplace's method which gives
\begin{align}
\lim_{d\to \infty} \frac{1}{d}\log Z= \max_{\ve \in [\ve_0,\,\ve_1]} \ \left[\alpha
  -I(\ve)-  \ve\right]\ .
  \label{eq:Laplace}
\end{align}

\begin{figure}
    \centering
    \includegraphics[width=0.45\textwidth]{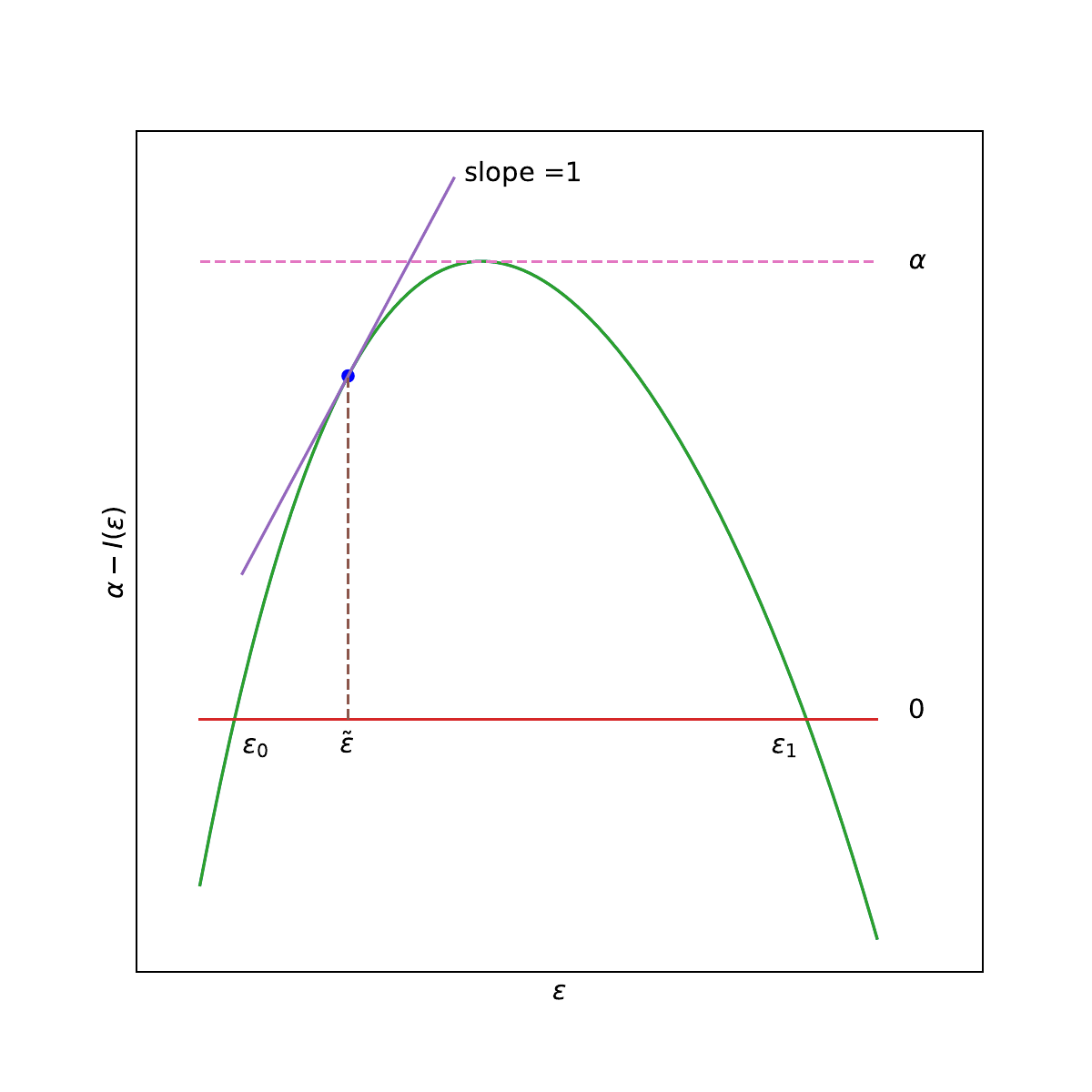}
     \includegraphics[width=0.45\textwidth]{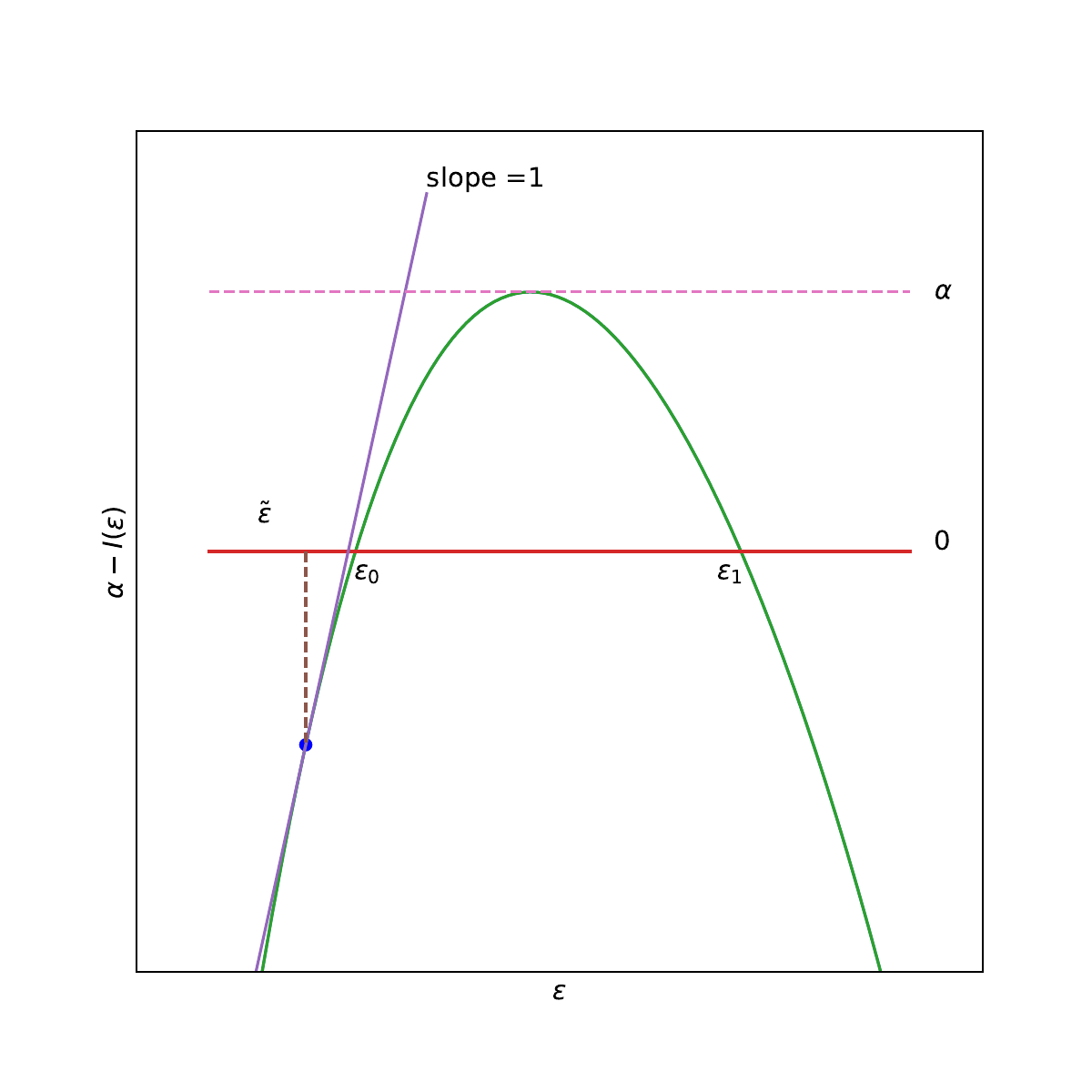}
    \caption{The REM phase transition. We sketch the behaviour of $\alpha-I(\ve)$ in the two phases. Left: the uncondensed phase; the slope at $\ve=\ve_0$ is larger than one, the integral in \ref{eq:REM_Zint} is dominated by the point $\tilde\ve$ which is inside the interval. Right :the condensed phase; the slope at $\ve=\ve_0$ is larger than one;  the integral in \ref{eq:REM_Zint} is dominated by the boundary $\ve=\ve_0$}
    \label{fig_REM}
\end{figure}

The location of the maximum depends on the value of the slope of the rate function at $\ve_0$ (see Fig.\ref{fig_REM}). Let us denote $\beta_c=-I'(\ve_0)$ (so that $\beta_c$ is a positive number which depends on $\alpha$ and on the distribution of energies). We have: :
\begin{enumerate}
    \item if $\beta_c>1$, the maximum in (\ref{eq:Laplace}) is obtained at $\ve=\tilde \ve>\ve_0$. The free entropy is given by $\phi=\alpha-I(\tilde\ve)-\tilde\ve$. This is called the uncondensed phase. 
    \item if $\beta_c<1$, the maximum in (\ref{eq:Laplace}) is obtained at $\ve =\ve_0$ and one finds $\phi=-\ve_0$ . This is called the condensed phase.
\end{enumerate}
The transition between these two regimes is called condensation transition. It takes place at a critical value of $\alpha=\log n/d$ defined by the solution of
\begin{align}
    |I'(\ve_0)|=1\ \ \text{where} \ \ \ve_0 \ \ \text{is the smallest root of }\ \alpha-I(\epsilon)=0
\end{align}
This critical value  $\alpha_c$ separates a phase $\alpha>\alpha_c$ which is uncondensed from a phase 
$\alpha<\alpha_c$ which is condensed.
 
In the condensed phase the partition function is dominated by the energy levels with the smallest possible energy, which is given by $\ve_0$. This can be studied as follows (see\cite{BouchaudMezard97,mezard2009information}). The distribution of the minimum $\ve_{min}$ of the energies $\ve_i\in \cS$ is a simple exercise in extreme event statistics. At large $d$, one finds that this distribution is peaked around $\ve_{min}\simeq \ve_0$, where $\ve_0$ is defined as before as the smallest $\ve$ such that
$I(\ve_0) =\alpha $. More precisely, if we write $\ve_{min}=\ve_0+ u/d$, then in the large $d$ limit the variable $u$ has a limiting distribution given by Gumbel's law: its probability density function is given by
\begin{align}
  \rho(u)= \beta_c\; e^{\beta_c u }\; \exp(-e^{\beta_c u} ) 
\end{align}
where $ \beta_c= |I'(\ve_0)|$.
If one focuses on the energy levels around $\ve_{min}$, one finds two important results \cite{BouchaudMezard97}:
\begin{itemize} 
\item The probability of the renormalized Boltzmann factors $z_i=e^{-\beta \ve_{i}}/e^{-\beta \ve_0}$, evaluated on the scale $z\sim O(1) $ follows a distribution which behaves as a power law:
\[
p(z)\sim \frac{1}{z^{\beta/\beta_c+1}} 
\]  
for $z\rightarrow \infty$ (but still of order one, i.e. large but not on a scale diverging with $d$).
\item  The Boltzmann weights $p_i=z_i/\sum_i z_i$ are distributed with a density \cite{BouchaudMezard97} $\rho(p)=(C/n) (1-p)^{\beta/\beta_c-1}p^{\beta_c/\beta+1}$
\end{itemize} 

Using this Gumbel distribution, it is possible to compute the participation ratios $Y_k=\mathbb{E} \sum_{i=1}^n p_i^k$. The computation, done in \cite{mezard1985random} and summarized in \cite{mezard2009information}, gives:
\begin{align}
  Y_k= 0\ \ \ \text{when}\ \  1<\beta_c \ \ \text{and}\ \  Y_k =\frac{\Gamma(k-\beta_c)}{\Gamma(k)\Gamma(\beta_c)}\ \ \ \text{when}\ \  \beta_c<1\nonumber\\
\end{align}
The fact that these ratios are finite in the large $d$ limit indicates that the partition function is dominated by the states with energies $\ve_i\simeq \ve_{min}$. In fact, one can show that the entropy density  $H=-\frac{1}{d}\sum_i p_i\log p_i$ vanishes in the large $d$ limit \cite{grossmezard1984}.

Let us now focus on the statistics of $Z_\cS/e^{-\beta \ve_0}$  for $\beta>\beta_c$.  In the condensed phase the partition function is dominated by the energy levels with the smallest possible energy,  whose associated renormalized Boltzmann factors are power-law distributed. In consequence,  $Z_\cS$ in the large d limit and in the condensed phase is a sum of i.i.d power law distributed random variables.  Using standard results on stable-laws \cite{nolan2012stable}, one can conclude that the probability distribution $Z_\cS/e^{-\beta \ve_0}$ follows an $\overline \alpha$-stable with $\overline \alpha=\beta_c/\beta$.  The renormalization by $e^{-\beta \ve_0}$ is needed to obtain a variable of order one in the asymptotic limit.  This result, first obtained in the physics literature \cite{mezard1987spin,gardner1989probability} has been put on a rigorous ground in \cite{bovier2002fluctuations} and extended in \cite{ben2005limit}.  
The reader will see the connection with regime (3) discussed in the main text.  

The case $1<\beta_c/\beta<2$ can be treated in a similar way \cite{gardner1989probability, bovier2002fluctuations,ben2005limit}.


\subsubsection{The REM replica free entropy}
The whole analysis of the REM above can be obtained using the REM replica free entropy defined as 
\begin{align}
    \Phi_{REM}(m)=\frac{1}{m}\alpha+\frac{1}{m}\overline{g}(m)
    \label{REM_free_ent}
\end{align}
where 
\begin{align}
\overline{g}(m) =\lim_{d\to\infty} \frac{1}{d} \log\left[ 
\int d \ve \; p_d(\ve) e^{-m d \ve}\right]\ .
\label{gbardef}
\end{align}
Note that this REM replica free-entropy differs by a factor $\alpha $ from the replica free entropy that we use for kernels, defined in \ref{def_repF}. The reason is that in the REM analysis one studies traditionally the partition function $Z_\cS$ which is a sum over $n$ terms, while in the kernel study the estimator involves $1/n$ times a sum over $n$ terms.

This replica free-entropy is originally found using the replica method an Parisi's one step RSB Ansatz\cite{mezard1987spin}. We shall not explain this approach here, but just check that all previosu results of the REM can be obtained through a study of the function $\Phi(m)$.

Using the large-deviation expression of the distribution of energies $p_d(\ve)$, the function $\overline g(m)$ can be computed by the Laplace method. The maximum of $-[I(\ve)+m\ve]$ is found at $\ve=\ove(m)$ which is the unique solution of 
\begin{align}
    I'(\ove(m))=-m\ ,
    \label{ovem_def}
\end{align}, and one obtains
$    \og(m)=-I(\ove(m))-m\ove(m)$. Using this expression and (\ref{ovem_def}), one obtains the simple expression:
\begin{align}
    \frac{d\Phi_{REM}}{dm}=\frac{1}{m^2}\left[
    I(\ove(m))-\alpha
    \right]
\end{align}
This gives $D=\frac{d\phi}{dm} (m=1)=I(\ove(1))-\alpha$.

As exemplified in Fig.\ref{fig_REM}, the case where $D<0$ corresponds to  $\ove(1)>\ve_0$, which is the uncondensed phase. In this case, the replica free entropy evaluated at $m=1$ gives $\Phi_{REM}(m=1)=\og(m=1)= \alpha-I(\ove(1))-\ove(1)$.

The second case, where $D>0$ corresponds to  $\ove(1)<\ve_0$, which is the condensed phase. Solving for 
$d\Phi_{REM}/dm=0$ gives a unique solution $m^*$ which is the solution of $I(\ove(m^*))=\alpha$. This implies that $\ove(m^*)=\ve_0$, and therefore $m^*=-I'(\ve_0)=\beta_c$ (notice that $0<m^*<1$).
The replica free entropy evaluated at $m=m^*$ gives 
\begin{align}
    \Phi_{REM}(m^*)&=
    \frac{1}{\beta_c}\alpha-\frac{1}{\beta_c}\left[   I(\ove(\beta_c))+\beta_c\ove(\beta_c)\right]\\
    &=-\ve_0
\end{align}

\subsubsection{Mapping the kernel density estimator to a REM }
Let us consider the kernel density estimator defined in (\ref{rhD_def}), for a given database $\mathcal {D}$ and a given point $x$, using a $\gamma$-kernel. We can write $\hat \rho_h^{\mathcal {D}}(x)=e^{d[-\alpha-\log h+c_\gamma]} Z_\cD$, where
$Z_\cD$ is a REM partition function, as defined in (\ref{Zremdef}), with energies
\begin{align}
   \ve_i= f\left(\frac{|x-y_i|^2}{d h^2}\right)\ .
    \label{epsilon_ker_def}
\end{align}
 One can also introduce the quadratic energies $u_i=\frac{|x-y_i|^2}{d h^2} $. We shall first study the distribution of the $u_i$, and then deduce the ones of the $\ve_i$ using the change of variable $\ve_i=f(u_i)$. 
 
 If $\rho$ is a pure density, for a given $x$,  the distribution of the $u_i$ variables is characterized by a cumulant-generating function $\tilde f_{d,x,h}(\lambda)$ which concentrates at large $d$ around its mean $\overline{f}_{h}(\lambda)=\lim_{d\to\infty}\int dx \rho(x) \tilde f_{d,x,h}(\lambda) $. 
Therefore the distribution of $u$
 satisfies a large deviation principle with a rate function $J_h(u)$ given by the Legendre transform:
 \begin{align}
     \overline{f}_{h}(\lambda)= -\min_u[\lambda u+ J_h(u)]
     \label{eq:Legendre}
 \end{align}
Therefore the distribution of the random energies satisfies a large deviation principle with a rate function $I_h(\ve)=J_h(f^{-1}(\ve))$. So the computation of $Z_{\cD}$ is exactly the one of the REM. As seen in the previous section, it can be done using the function $\overline g $ defined in (\ref{gbardef}). In our case, this function depends on the parameter $m$ and the bandwidth $h$, and it is given by (\ref{gbarh2m}). Therefore the replica free-entropy defined in (\ref{free_ent_def}) is identical to the one found in the study of the REM (see(\ref{REM_free_ent})).

 \subsubsection{Proof of Lemma \ref{lemma_hc}}
For the $\gamma$-kernels defined in (\ref{gammaKerdef}), 
one has $\overline g(h^2,m)=m c_\gamma -m \log h+G(m/h^{2\gamma})$,
with
\begin{align}
\label{Gdef}
    G(u)=\frac{1}{d}\; \mathbb{E}_x\; \log \left[\mathbb{E}_y \; \exp\left(-u\frac{d}{2\gamma}[(x-y)^2/d]^\gamma\right)\right]
\end{align}
so that 
\begin{align}
    \phi_{\alpha,h}(m)= +c_\gamma-\log h+\frac{1}{m}
    \left(\alpha +G\left( \frac{m}{h^{2\gamma}}\right)\right)\ .
\end{align}
The derivative of $\phi$ with respect to $m$ is
\begin{align}
    \frac{\partial \phi_{\alpha,h}(m)}{\partial m}=\frac{1}{m^2}\left[
    -\alpha
     -G\left(\frac{m}{h^{2\gamma}}\right)+ \frac{m}{h^{2\gamma}}G'\left(\frac{m}{h^{2\gamma}}\right) 
     \right]
\end{align}
As $G$ is a convex function with positive second derivative on $\mathbb{R}^+$, the function $m^2 d\phi_{\alpha,h}/dm$ is a monotonously increasing function of $m/h^{2\gamma}$ 
Using the convexity of $G$, we see that $H(x)=xG'(x)-G(x)$ is an increasing function of its argument.
Its derivative at $m=1$, $D(\alpha,h)$, is a  decreasing function of $\alpha$ and a decreasing  function of $h^2 $. 
When $h\to\infty$, one finds $D(\alpha,h)\sim-\alpha<0$. When $h\to 0$, the integral in the definition of $g(x,h^2,m)$ is dominated by $a\sim x +O(h)$ and therefore $g(x,h^2,m)$ diverges as $(1-m) \log h$. From this behavior one deduces that $D(\alpha,h)\to+\infty$. 
Therefore for a fixed $\alpha$, the equation in $h$, $D(\alpha,h)=0$, has a unique solution $h_G(\alpha)$.
 $\square$
\subsubsection{Proof of Theorem \ref{thm_concentration_pure}}
Lemma\ref{lemma_hc} shows the existence of a critical value of the bandwidth, $h_G(\alpha)$. For $h>h_G(\alpha)$,  $D(\alpha,h)<0$ and the replica free entropy analysis shows that the REM defined by $Z_{\cD}$ is uncondensed. In this phase, $\frac{1}{d}\log Z_{\cD}$ concentrates at large $d$ towards $\frac{1}{d}\log{\mathbb{E}}_{\cD} Z_{\cD}$. This gives the expression (\ref{free_large_h}). For 
$h<h_G(\alpha)$,  $D(\alpha,h)>0$ and the replica free entropy analysis shows that the REM defined by $Z_{\cD}$ is condensed. Then one must find the value of $m^*$ such that $d\phi_{\alpha,h}/dm \;(m^*)=0$. One can prove that this value is unique by the following reasoning:  $m^2 d\phi_{\alpha,h}/dm$ is a monotonously increasing function of $m$; therefore $m^2d/dm[m^2 d\phi_{\alpha,h}/dm]>0$, which shows that $\phi_{\alpha,h} $ is a convex function of $m$;  
 when $m\to 0$, we have
 $m^2 d\phi_{\alpha,h}/dm=-\alpha $ ,  $d\phi_{\alpha,h}/dm$ is an increasing function of $m$, it goes to $-\infty$ when $m\to 0$ and it goes to a positive value when $m=1$, therefore there exists a unique $m^* \in (0,1)$ where $d\phi_{\alpha,h}/dm=0$. 
 Then $\frac{1}{d}\log Z_{\cD}$ concentrates at large $d$ towards $\Phi(m^*)$. This gives the expression (\ref{free_large_h}).
 $\square$

 \subsubsection{Proof of Corollary \ref{coroll_DKL_pure}}
The first part of Corollary \ref{coroll_DKL_pure} - the relation between KL divergence and replica free entropy - is a direct consequence of the expression of $\hat \rho_h$ found in Theorem \ref{thm_concentration_pure}. Now we obtain the condition on the rate function stated in Corollary \ref{coroll_DKL_pure}. 

Let us focus on the uncondensed phase 
$h>h_{G}$. We can express $\overline g (h^2,m=1)$
using the variable $u= |x-y|^2/d$ which satisfies a large deviation principle with a rate function $J_1(u)$ defined in (\ref{eq:Legendre}). We find
\begin{align}
    \overline g (h^2,m=1)=-\log h+c_\gamma -J_1(u^*)-\frac{1}{2\gamma} \left(\frac{u^*}{h^2}\right)^\gamma
\end{align}
and therefore 
\begin{align}
D_{KL}(\rho||\hat \rho_h)=J_1(u^*)+\frac{1}{2\gamma}\left(\frac{u^*}{h^2}\right)^\gamma+\log h + c
\end{align}
where $c$ contains terms independent of $h$, and $u^*$ is the value of $u$ where the right-hand side reaches its maximum; it satisfies the equation:
\begin{align}
    \frac{dJ_1(u^*)}{du}=-\frac{1}{2}\left(\frac{u^*}{h^2}\right)^{\gamma-1}\frac{1}{h^2}
\end{align}
Using these two equations one finds
\begin{align}
\frac{dD_{KL}(\rho||\hat \rho_h)}{dh}=\frac{1}{h}
\left(1+2 u^* \frac{dJ_1(u^*)}{du}
\right)
\end{align}
which is strictly positive if $2 \frac{dJ_1(u^*)}{du}u^*>-1$. 

For the $\gamma$-kernels we consider, $u^*$ is an increasing function of $h$. 
 Therefore, requiring that $2 \frac{dJ_1(u^*)}{du}u^*>-1$ for all $h>h_{CLT}$ is equivalent to requiring 
$2 \frac{dJ_1(u)}{du}u>-1$ for $u_G<u<u_{typ}$, where $u_{G}=u^*(h_{CLT})$ and $u_{typ}=\lim_{h\rightarrow +\infty}u^*(h)$ is the value of $u$ where $J_1$ vanishes. The equation on $u^*(h)$ above, implies that $\frac{dJ_1(u^*)}{du}=0$
for $u_{typ}$. Repeating this procedure for $\phi_{\alpha,h}$, one finds that the condition for $u_G$ is $J_1(u_G)=-\alpha$. Therefore, if $2 \frac{dJ_1(u)}{du}u>-1$ for $u_G<u<u_{typ}$, the derivative of the KL divergence is strictly positive in the RS phase ($h>h_{CLT}$). Furthermore, it is easy to show that the derivative of the KL divergence is negative for $h$ small enough. In consequence, if the required condition above is satisfied the minimum of the KL divergence takes place for $h\le h_{CLT}$.   
$\square$
 $\square$

\subsubsection{ Corollaries \ref{thm_gauss} and \ref{thm_alpha} }
 Having proven the main Theorem \ref{thm_concentration_pure}, the corollaries \ref{thm_gauss} and \ref{thm_alpha} can be obtained by straightforward extension of the proofs developed in \cite{bovier2002fluctuations,ben2005limit}. We shall not develop them here.

\section{Analysis of Gaussian densities}
\label{GaussianDensity_proofs}
\subsection{Concentration: Proof of Proposition \ref{prop_Concentration_Gauss} }
\label{proof_Concentration_Gauss}
Consider a variable $ y \in{\mathbb{R}}^d $ generated from the Gaussian density defined in Sec.(\ref{sec:gaussexa}), with covariance matrix $C$. At fixed $x$, let us study
the distribution $\rho_x(l)$ of the random variable $l$ defined by
$
    l= (x-a)^2/d= (1/d) \sum_i(x_i-y_i)^2
    $.

It is easy to compute the logarithmic generating function of its moments, defined for $t\geq 0$ as:
\begin{align}
\psi_x(t)\equiv\frac{1}{d}\log\left[\int dl \rho_x(l) e^{-t d l/2} \right]
=-\frac{1}{2d}\text{Tr} \log (1+t C)-\frac{t}{2d}
\; x^T (1+tC)^{-1} x
\end{align}
When $x$ is generated from the distribution $\rho$, this concentrates at large $d$ to 
\begin{align}
\psi(t)=\lim_{d\to \infty} \int dx \rho (x)\psi_x(t)=- \frac{1}{2} \int d\lambda \rho_{C}(\lambda)\left[
 \log (1+t \lambda)+ \frac{t\lambda}{1+t\lambda}
\right]
\end{align}
We notice that this function is concave and twice differentiable  for all positive $t$. Using the Gärtner-Ellis theorem, we can deduce that the probability density of $l$, $\rho_x(l)$, evaluated at a generic point $x$ sampled from $\rho$, satisfies a large deviation principle
\begin{align}
\lim_{d\to\infty} \int dx \rho(x) \frac{1}{d}\log \rho_x(l)   =-I(l) 
\end{align}
where $I(l)$ and $\psi(t)$ are related by a Legendre transformation:
$
    I(l)=\text{Sup}_t\left[-\psi(t)-\frac{t l}{2}\right]
$.

We now consider the function
\begin{align}
    g(x,h^2,m)&=\frac{1}{d} \log \int \frac{d^d y}{h^{md}}\; \rho(y) \; e^{d m\left(c_\gamma- 
    f_\gamma\left(\frac{(x-y)^2}{d h^2}\right)\right)}\\
    &= \frac{1}{d} \log\int \frac{dl}{h^{md}} \rho_x(l)\; e^{d m\left(c_\gamma- 
    f_\gamma\left(\frac{l}{ h^2}\right)\right)}
\end{align}
When $x$ is distributed from $\rho$, this concentrates to
\begin{align}    
    \overline g(h^2,m) &= mc_\gamma-m\log h +\text{Sup}_l    \left(-I(l)-m f_\gamma\left(\frac{l}{h^2}\right)\right)\\
 &= mc_\gamma-m\log h +\text{Sup}_l \text{Inf}_t\left(\psi(t)+\frac{t l}{2}-m f_\gamma\left(\frac{l}{h^2}\right)\right)
\end{align}
In our case,  $- \psi$ and $f_\gamma$ are concave and twice differentiable everywhere. Therefore there is a unique value of the pair $l,t$ where the $\text{Sup}_l \text{Inf}_t$ is found. This pair is found by the stationnarity condition
\begin{align} 
\frac{t}{2}=\frac{m}{h^2}f_\gamma'\left( \frac{l}{h^2}\right)\ \ ;\ \ \frac{l}{2}+\frac{d\psi}{dt}=0
\end{align}
which gives  
Proposition\ref{prop_Concentration_Gauss} (with $l, \hat l$ denoting the value of $l,t$ at stationnarity). $\square$

\subsection{Critical line: Proof of Proposition \ref{prop_CL_Gauss}}
\label{proof_CL_Gauss}
$D(\alpha,h)$ is a decreasing function of $\alpha$ at fixed $h$. Let us show that it is a decreasing function of $h$ at fixed $\alpha$. 
 
We write the two equations (\ref{lhatl}) as $\hat l=m u f'(u l)$ and $l=g(\hat l)$, where $u=1/h^2$ and $g(x)$ is a monotonously decreasing function. Then one has 
$\hat l^*=m u f'(u g(\hat l^*))$. From this one deduces 
\begin{align}
    \frac{\partial l^*}{\partial u}=m  f'(u g(\hat l^*))+m u f''[u g(\hat l^*)] \left[g(\hat l^*)+ug'(\hat l^*)\frac{\partial l^*}{\partial u}\right]
\end{align}
and using the positivity of $f',f''$ and $-g'$ one deduces that $\frac{\partial l^*}{\partial u}>0$, and thus  $\partial \hat l^*/\partial h<0 $
Similarly,
\begin{align}
    \frac{\partial l^*}{\partial m}=u f'(u g(\hat l^*))+m u f''[u g(\hat l^*)]\;  ug'(\hat l^*)\;\frac{\partial l^*}{\partial m}
\end{align}
and using the positivity of $f',f'' and -g'$ one deduces that $\frac{\partial l^*}{\partial m}>0$

We have shown that $\partial \hat l^*/\partial m>0 $ and $\partial \hat l^*/\partial h<0 $. As
$D$ is an increasing function of $\hat l^*$, we have  $\partial D/\partial h<0$.
  
One easily sees that: for fixed $h$, $D$ is positive at small $\alpha$ and goes to $-\infty$ at large $\alpha$ ; for fixed $\alpha$, $D$ goes to $+\infty$ at small $h$ and goes to $-\alpha $ at large $h$. 
Together with the monotonicity property that we have just established, it shows that the equation $D=0 $ has a unique solution in $h$.  $\square$

\subsection{Kullback-Leibler divergence and optimal value of $h$: Proof of Proposition \ref{prop_KL_Gauss} }
\label{proof_KL_Gauss}
\subsubsection{RS phase}
We first study the RS expression $\phi^{RS}_{\alpha}(h^2)=
\phi_{\alpha,h}(m=1)$ where $\phi_{\alpha,h}(m) $ is given in
(\ref{phi_Gauss}) and $l,\hat l$ are the solutions of Eqs.(\ref{lhatl})
with $m=1$. We shall show that $\frac{d\phi^{RS}_{\alpha}(h^2)}{d
  h^2}<0$. We start from
\begin{align}
  \frac{d\phi^{RS}_{\alpha}(h^2)}{d h^2}=
  -\frac{1}{2 h^2}+\frac{l_1}{h^4}f'\left(\frac{l_1}{h^2} \right)
  \end{align}
  Using the fact that $l_1,\hat l_1$ are the solutions of 
  \begin{align}
    \frac{\hat l_1}{2}&=\frac{1}{h^2}f'\left(\frac{l_1}{h^2}\right) \label{lhatl1}\\
    l_1&=\int d\lambda \rho_C(\lambda) \left(\frac{\lambda}{1+\hat l_1 \lambda}+\frac{\lambda}{(1+\hat l_1 \lambda)^2}\right)\label{lhatl2}
  \end{align}
  we obtain
  \begin{align}
  \frac{d\phi^{RS}_{\alpha}(h^2)}{d h^2}=
    \frac{1}{2 h^2}( l_1 \hat l_1-1)
    = - \int d\lambda \rho_C(\lambda) \left(\frac{1}{(1+\hat l_1 \lambda)^2}
    \right)
  \end{align}
  which is negative. From Corollary 7, we therefore see that in
  the whole  RS phase, the Kullback-Leibler divergence is an increasing
  function of $h$, therefore it is minimum at the RS-1RSB phase
  transition $h=h_G(\alpha)$.  Note that for the multivariate Gaussian distribution one has $2\frac{dJ_1(u)}{du}u=1-\frac{u}{u_{typ}}$, which indeed verifies the condition of Corollary 7. 
  \subsubsection{1RSB phase}
  We now study the 1RSB phase,where  $\phi^{1RSB}_{\alpha}(h^2)=
\phi_{\alpha,h}(m^*)$ where $\phi_{\alpha,h}(m) $ is given in
(\ref{phi_Gauss}), $l,\hat l$ are the solutions of
Eqs.(\ref{lhatl}), and $m$ is fixed to the value  where
$d\phi_{\alpha,h}(m)/dm =0$. We now find:
 \begin{align}
  \frac{d\phi^{1RSB}_{\alpha}(h^2)}{d h^2}=
   \frac{1}{2 h^2}\left( \frac{l \hat l}{m}-1\right)
 \end{align}
 where
 $m,l,\hat l$ are the solutions of the three equations:
 \begin{align}
   \alpha&=\frac{1}{2}\int d\lambda \rho_C(\lambda) \left(
\log(1+\lambda\hat l)-\frac{\lambda\hat l}{(1+\lambda\hat l)^2}
   \right) \label{eqa}\\
     l&=\int d\lambda \rho_C(\lambda ) \left(\frac{\lambda}{1+\hat
          l \lambda}+\frac{\lambda}{(1+\hat l
          \lambda)^2}\right)\label{eqb}\\
   m&=  \frac{\hat l h^2}{2 f'\left(\frac{l}{h^2}\right)} \label{eqc}
  \end{align}
    Given $\alpha$ and the spectral distribution of the Gaussian
    density $\rho_C$, Eq.(\ref{eqa}) is easily solved for $\hat l$, as
    the right-hand side is an increasing function of $\hat l$. Then,
    Eq.(\ref{eqb}) gives $l$, and Eq.(\ref{eqc}) gives the value of
    $m$.

    From Eq.(\ref{eqc}) one sees that $m$ is an increasing function of
    $h^2$, which vanishes when $h\to 0$. We
    also know that $m=1$ on the critical line
    $h=h_G(\alpha)$. Therefore $l \hat l /m -1$ vanishes at a unique
    value $h=h^*(\alpha)$ which is in the 1RSB phase:
    $h^*(\alpha)<h_G(\alpha)$. From Eq.(\ref{eqc}) one finds that $h$ satisfies $ (l/h^2) f'(l/h^2)=1/2$. $\square$
    
\subsection{Proof of Proposition \ref{prop_KL_opt_Gauss}}
 We can now use the general formula (\ref{DKL_1RSB}) to compute the KL divergence. We find
   \begin{align}
       D_{KL}= -\frac{1}{2} \log(2\pi e)-\frac{1}{2}\langle \lambda\rangle 
       -\frac{1-m}{m}\alpha +\log h+
       \frac{1}{2m}\langle \log(1+\hat l \lambda)\rangle
+\frac{1}{2m}\langle \frac{\hat l\lambda}{1+\hat l\lambda}\rangle
-\frac{\hat l l}{2m}+f\left(\frac{l}{h^2}\right)
   \end{align}
It is interesting to note that this formula is fully variational: The four equations $dD_{KL}/dl=0,dD_{KL}/d\hat l=0,dD_{KL}/dm=0,dD_{KL}/dh=0$ give back the equations (\ref{eqa},\ref{eqb},\ref{eqc}) and the optimality condition for $h$ ($l \hat l =m$).
This expression for the minimal KL divergence can be simplified using the explicit expression for $c_\gamma$ given in (\ref{cgammadef}), leading to the expression of Proposition \ref{prop_KL_Gauss}. $\square$

\section*{Acknowledgement}

We thank F. Bach, A. Lijoi, A. Montanari and M. Wainwright for discussions. GB acknowledges
support from the ANR PRAIRIE. MM acknowledges financial support by the PNRR-PE-AI FAIR
project funded by the NextGeneration EU program.

 \bibliographystyle{plain}

\end{document}